\documentclass{article}

\usepackage{microtype}
\usepackage{graphicx}
\usepackage{subfigure}
\usepackage{booktabs} % for professional tables

\usepackage{hyperref}

\usepackage[accepted]{icml2023}

\usepackage{amsmath}
\usepackage{amssymb}
\usepackage{mathtools}
\usepackage{amsthm}

\usepackage[capitalize,noabbrev]{cleveref}

\usepackage{multirow}
\usepackage{makecell}
\usepackage{pgfplots}
\pgfplotsset{compat=newest}
\pgfplotsset{scaled y ticks=false}
\usepgfplotslibrary{groupplots}
\usepgfplotslibrary{dateplot}
\pgfplotsset{compat=1.11,
	every axis plot/.append style={line width=1pt},
	legend style={font=\tiny}
}

\usepackage{tikz}
\usetikzlibrary{plotmarks}

\newcommand{\R}{\mathbb{R}}
\newcommand{\J}{\mathbb{J}}
\newcommand{\C}{\mathbb{C}}
\newcommand{\D}{\mathbb{D}}
\DeclareMathOperator{\diag}{diag}
\DeclareMathOperator{\cayley}{Cayley}
\newcommand{\FFT}{\mathsf{FFT}}
\newcommand{\iFFT}{\mathsf{FFT}^{-1}}
\newcommand{\hin}{h_{\mathsf{in}}}
\newcommand{\hout}{h_{\mathsf{out}}}
\newcommand{\vecc}{\operatorname{Vec}}

\theoremstyle{plain}
\newtheorem{theorem}{Theorem}[section]
\newtheorem{proposition}[theorem]{Proposition}
\newtheorem{lemma}[theorem]{Lemma}

\theoremstyle{definition}
\newtheorem{definition}[theorem]{Definition}
\newtheorem{assumption}[theorem]{Assumption}
\theoremstyle{remark}
\newtheorem{remark}[theorem]{Remark}
\newtheorem{example}[theorem]{Example}

\usepackage[textsize=tiny]{todonotes}

\icmltitlerunning{Lipschitz-Bounded Deep Networks\hfill\thepage}
\begin{document}

\twocolumn[
\icmltitle{Direct Parameterization of Lipschitz-Bounded Deep Networks}




\begin{icmlauthorlist}
\icmlauthor{Ruigang Wang}{acfr}
\icmlauthor{Ian R. Manchester}{acfr}
\end{icmlauthorlist}

\icmlaffiliation{acfr}{Australian Centre for Robotics, School of Aerospace, Mechanical and Mechatronic Engineering, The University of Sydney, Sydney, NSW 2006, Australia}

\icmlcorrespondingauthor{Ruigang Wang}{ruigang.wang@sydney.edu.au}

\icmlkeywords{Machine Learning, ICML}

\vskip 0.3in
]



\printAffiliationsAndNotice{}  % leave blank if no need to mention equal contribution

\begin{abstract}
This paper introduces a new parameterization of deep neural networks (both fully-connected and convolutional) with guaranteed $\ell^2$ Lipschitz bounds, i.e. limited sensitivity to input perturbations. The Lipschitz guarantees are equivalent to the tightest-known bounds based on certification via a semidefinite program (SDP). We provide a ``direct'' parameterization, i.e., a smooth mapping from $\mathbb R^N$ onto the set of weights satisfying the SDP-based bound. Moreover, our parameterization is  complete, i.e. a neural network satisfies the SDP bound if and only if it can be represented via our parameterization. This enables training using standard gradient methods, without any inner approximation or computationally intensive tasks (e.g. projections or barrier terms) for the SDP constraint. The new parameterization can equivalently be thought of as either a new layer type (the \textit{sandwich layer}), or a novel parameterization  of standard feedforward networks with parameter sharing  between neighbouring layers. A comprehensive set of experiments on image classification shows that sandwich layers outperform previous approaches on both empirical and certified robust accuracy. Code is available at \url{https://github.com/acfr/LBDN}.
\end{abstract}

%%%%%%%%%%%%%%%%%%%%%%%%%%%%%%%%%%%%%%%%%%%%%%%%%%%%%%%%%%%%%%%%%%%%%%%%%%%%%%%
\section{Introduction}

Neural networks have enjoyed wide application due to their many favourable properties, including highly accurate fits to training data, surprising generalisation performance within a distribution, as well as scalability to very large models and data sets. Nevertheless, it has also been observed that they can be highly sensitive to small input perturbations  \cite{szegedy2013intriguing}. This is a critical limitation in applications in which certifiable robustness is required, or the smoothness of a learned function is important.

A standard way to quantify sensitivity of models is via a \textit{Lipschitz bound}, which generalises the notion of a slope-restricted scalar function. A function $x\mapsto f(x)$ between normed spaces satisfies a Lipschitz bound $\gamma$ if 
\begin{equation}\label{eq:Lip}
\|f(x_1)-f(x_2)\|\leq \gamma \|x_1-x_2\|
\end{equation}
for all $x_1, x_2$ in its domain. The (true) Lipschitz constant of a function, denoted by $\mathrm{Lip}(f)$, is the smallest such $\gamma$. Moreover, we call $f$ a \emph{$\gamma$-Lipschitz} function if $\mathrm{Lip}(f)\leq \gamma$.

A natural application of Lipschitz-bounds is to control a model's sensitivity to \textit{adversarial} (worst-case) inputs, e.g. \cite{madrytowards, tsuzuku2018lipschitz}, but they can also be effective for regularisation \cite{gouk2021regularisation} and Lipschitz constants often appear in theoretical generalization bounds \cite{Bartlett:2017,bubeck2023universal}. Lipschitz-bounded networks have found many applications, including: stabilising the learning of generative adversarial networks \cite{arjovsky2017wasserstein, gulrajani2017improved}; implicit geometry mechanisms for computer graphics \cite{liu2022learning}; in reinforcement learning to controlling sensitivity to measurement noise (e.g. \cite{russo2021towards}) and to ensure robust stability of feedback loops during learning \cite{wang2022youla}; and the training of differentially-private neural networks \cite{bethune2023dp}. In robotics applications, several learning-based planning and control algorithms require known Lipschitz bounds in learned stability certificates, see e.g. the recent surveys \cite{brunke2022safe, dawson2023safe}.

Unfortunately, even for two-layer perceptrons with ReLU activations, exact calculation of the true Lipschitz constant for $\ell^2$ (Euclidean) norms is NP-hard \cite{virmaux2018lipschitz}, so attention has focused on approximations that balance accuracy with computational tractability. Crude $\ell^2$-bounds can be found via the product of spectral norms of layer weights \cite{szegedy2013intriguing}, however to date the most accurate polynomial-time computable bounds require solution of a semidefinite program (SDP) \cite{fazlyab2019efficient}, which is computationally tractable only for relatively small fully-connected networks. 

While \textit{certification} of a Lipschitz bound of a network is a (convex) SDP with this method, the set of weights satisfying  a prescribed Lipschitz bound is non-convex, complicating training. Both \cite{rosca2020case} and \cite{dawson2023safe} highlight the computationally-intensive nature of SDP-based bounds as limitations for applications. 

\paragraph{Contribution.} In this paper we introduce a new parameterization of standard feedforward neural networks, both fully-connected multi-layer perceptron (MLP) and deep convolutional neural networks (CNN).
\begin{itemize}
    \item The proposed parameterization has \textit{built-in} guarantees on the network's Lipschitz bound, equivalent to the best-known bounds provided by the SDP method \cite{fazlyab2019efficient}.
    \item Our parameterization is a smooth surjective mapping from an unconstrained parameter space $\R^N$ onto the (non-convex) set of network weights satisfying these SDP-based bounds. This enables learning of lipschitz-bounded networks via standard gradient methods, avoiding the complex projection steps or barrier function computations that have previously been required and limited scalability.
    \item The new parameterization can equivalently be treated as either a composition of new $1$-Lipschitz layers called \emph{Sandwich} layer, or a parameterization of standard feedforward networks with coupling parameters between neighbouring layers. 
\end{itemize}

\paragraph{Notation.}  Let $\R$ be the set of real numbers. $A\succeq 0$ means that a square matrix $A$ is a positive semi-definite. We denote by $\D_{++}^n$ for the set of $n\times n$ positive diagonal matrices. For a vector $x\in \R^n$, its $2$-norm is denoted by $\|x\|$. Given a matrix $A\in \R^{m\times n}$, $\|A\|$ is defined as its the largest singular value and $A^+$ denotes its Moore–Penrose pseudoinverse.

%%%%%%%%%%%%%%%%%%%%%%%%%%%%%%%%%%%%%%%%%%%%%%%%%%%%%%%%%%%%%%%%%%%%%%%%%%%%%%%
\section{Problem Setup and Preliminaries}

Consider an $L$-layer feedforward neural network $y=f(x)$ described by the following recursive equations:
\begin{equation}\label{eq:DNN}
      \begin{split}
        z_0&=x,\\
	z_{k+1} &=\sigma(W_k z_k+b_k),\quad k=0,\ldots,L-1 \\
	y&=W_L z_L+b_L,
  \end{split}
\end{equation}
where $x\in \R^{n_0}, z_k \in \R^{n_k}, y\in \R^{n_{L+1}}$ are the network input, hidden unit of the $k$th layer and network output, respectively. Here $W_k \in \R^{n_{k+1}\times n_k}$ and $b_k\in \R^{n_{k+1}}$ are the weight matrix and bias vector for the $k$th layer, and $\sigma$ is a scalar activation function applied element-wise. We make the following assumption, which holds for most commonly-used activation functions (possibly after rescaling) \cite{goodfellow2016deep}.
\begin{assumption}\label{asmp:sigma}
    The nonlinear activation $\sigma:\R\rightarrow\R$ is piecewise differentiable and slope-restricted in $[0,1]$.
\end{assumption}
If different channels have different activation functions, then we simply require that they all satisfy the above assumption. 

The main goal of this work is to learn feedforward networks \eqref{eq:DNN} with certified Lipschitz bound of $\gamma$, i.e.,
\begin{equation}\label{eq:robust-ml}
    \min_{\phi}\; \mathcal{L}(f_\phi)\quad \mathrm{s.t.} \quad \mathrm{Lip}(f_\phi)\leq \gamma
\end{equation}
where $\mathcal{L}(\cdot)$ is a loss function and $\phi:=\{(W_k,b_k)\}_{0\leq k \leq L}$ is the learnable parameter. Since it is NP-hard to compute $\mathrm{Lip}(f_\phi)$, we seek an accurate Lipschitz bound estimation so that the constraint in \eqref{eq:robust-ml} does not lead to a significant restriction on the model expressivity. 

In \cite{fazlyab2019efficient}, integral quadratic constraint (IQC) methods were applied to capture both monotonicity and 1-Lipschitzness properties of $\sigma$, leading to state-of-the-art Lipschitz bound estimation based on the following linear matrix inequality (LMI), see details in \cref{sec:lmi}: 
\begin{equation}\label{eq:lmi}
    H:=\begin{bmatrix}
        \gamma I & -U^\top \Lambda & 0\\
        -\Lambda U & 2\Lambda-\Lambda W-W^\top \Lambda & -Y^\top \\
        0 & -Y & \gamma I
    \end{bmatrix}\succeq 0
\end{equation}
where $\Lambda \in \D_{++}^n$ with $n=\sum_{k=1}^{L}n_k$, and
\begin{gather*}
    W=\begin{bmatrix}
    0 & & &\\
    W_1 & \ddots & & \\
    \vdots & \ddots & 0 & \\
    0 & \cdots & W_{L-1} & 0 
  \end{bmatrix}, \;
  U=\begin{bmatrix}
     W_0 \\ 0 \\ \vdots \\ 0
  \end{bmatrix}, \\
  Y=\begin{bmatrix}
    0 & \cdots & 0 & W_L
  \end{bmatrix}.
\end{gather*}

\begin{remark} The published paper \cite{fazlyab2019efficient} claimed that even tighter Lipschitz bounds could be achieved with a less restrictive class of multipliers $\Lambda$ than diagonal. However, this is false: a counterexample was presented in \citep{pauli2021training}, and the error in the proof was explained in \cite{revay2020lipschitz}, see also Remark \ref{rem:error} in the appendix of this paper.
\end{remark}

In this paper we approach problem \eqref{eq:robust-ml} via model parameterizations guaranteeing a given Lipschitz bound.
\begin{definition}
    A parameterization of the network \eqref{eq:DNN} is a differentiable mapping $\phi=\mathcal{M}(\theta)$ where $\theta \in \Theta \subseteq \R^N$. It is called a \emph{convex parameterization} if $\Theta$ is convex, and a \emph{direct parameterization} if $\Theta=\R^{N}$. 
\end{definition}

Given a network with fixed $W,U,Y$, Condition \eqref{eq:lmi} is  convex  with respect to the Lipschitz bound $\gamma$ and multiplier $\Lambda$. When training a network with specified bound $\gamma$, we can convert \eqref{eq:lmi} into a convex parameterization by introducing new decision variables $\tilde U= \Lambda U,\tilde W=\Lambda W$, \eqref{eq:lmi} becomes
\begin{equation}\label{eq:sparse-H}
    H=\begin{bmatrix}
    \gamma I & - \hat{W}_0^\top & \\
    -\hat{W}_0 & 2\Lambda_0  & -\hat{W}_1^\top \\
    & \ddots & \ddots & \ddots \\
    & & -\hat W_{L-1} & 2\Lambda_{L-1} & -\hat W_L^\top \\
    & & & -\hat W_L & \gamma I
    \end{bmatrix}\succeq 0
\end{equation}
where $\hat{W}_k=\Lambda_k W_k$ for $0\leq k<L$ and $\hat{W}_L=W_L$. In \cite{pauli2021training,revay2020convex}, constrained optimization methods such as convex projection and barrier functions are applied for training. However, even for relatively small-scale networks (e.g. $ \sim1000$ neurons), the associated barrier terms or projections become a major computational bottleneck.  

%%%%%%%%%%%%%%%%%%%%%%%%%%%%%%%%%%%%%%%%%%%%%%%%%%%%%%%%%%%%%%%%%%%%%%%%%%%%%%%
\section{Direct Parameterization}

In this section we will present our main contribution -- a direct parameterization $\phi=\mathcal{M}(\theta)$ such that $\phi$ automatically satisfies \eqref{eq:lmi} and hence \eqref{eq:Lip} for any $\theta \in \R^N$. Then, the learning problem \eqref{eq:robust-ml} can be transformed into an unconstrained optimization problem
\[
\min_{\theta\in \R^N}\quad \mathcal{L}(f_{\mathcal{M}(\theta)}).
\]

\begin{figure}[t]
\vskip 0.2in
\begin{center}
\includegraphics[width=\columnwidth]{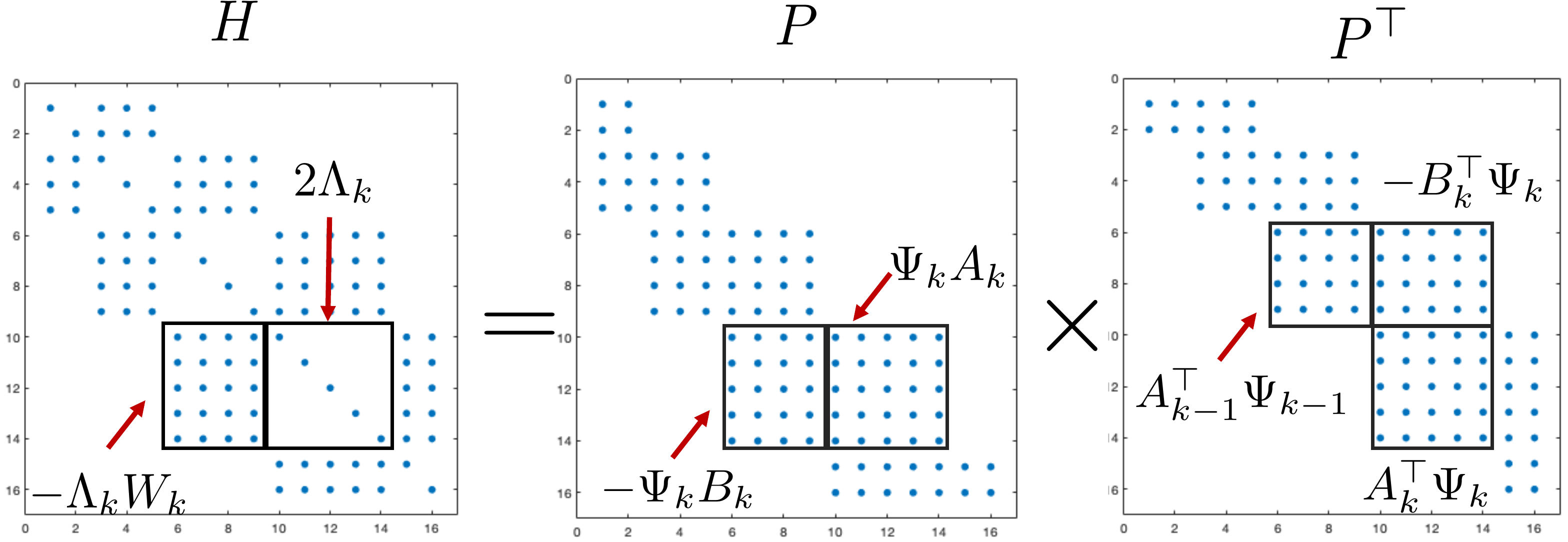}
\caption{Illustration of the sparsity pattern in $H$ that must be preserved by the factorization $H=PP^\top$.}
\label{fig:H}
\end{center}
\vskip -0.2in
\end{figure}

First, it is clear that \eqref{eq:sparse-H} is satisfied if we parameterize $H$ as $H=PP^\top$. The main challenge then is to parameterize $P$ such that the particular sparsity pattern of $H$ is recovered: a block-tridiagonal structure where the main diagonal blocks must be positive diagonal matrices, see \cref{eq:sparse-H} and \cref{fig:H}. First, the block-tridiagonal structure can be achieved by taking
\[
P=
\begin{bmatrix}
    D_{-1} &\\ 
    -V_0 & D_0 &\\
    & \ddots & \ddots & \\
    & & -V_L & D_L 
\end{bmatrix},
\]
i.e., block Cholesky factorization of $H$. The next step is to construct $V_k$ and $D_k$ such that the diagonal blocks $H_{kk}=V_kV_k^\top +D_k D_k^\top$ are diagonal matrices. We do so by the Cayley transform for orthogonal matrix parameterization \cite{trockman2021orthogonalizing}, i.e., for any $X\in \R^{m\times m}, Y\in \R^{n\times m}$ we have $Q^\top Q=I$ if
\begin{equation}
    Q=\cayley\left(\begin{bmatrix}
        X \\ Y
    \end{bmatrix}\right):=\begin{bmatrix}
        (I+Z)^{-1}(I-Z) \\
        -2Y(I+Z)^{-1}
    \end{bmatrix} \label{eq:cayley}
\end{equation}
with $Z=X-X^\top + Y^\top Y$. To be more specific, we take $D_k=\Psi_k A_k$ and $ V_k=\Psi_k B_k$  where 
\begin{equation}\label{eq:PAB}
    \Psi_k =\diag\bigl(e^{d_k}\bigr),\quad \begin{bmatrix}
        A_k^\top \\ B_k^\top
    \end{bmatrix} =\cayley\left(\begin{bmatrix}
        X_k \\ Y_k
    \end{bmatrix}\right)
\end{equation}
for some free vector $d_k$ and matrices $X_k,Y_k$ with proper dimension. Now we can verify that $H=PP^\top$ has the same structure as \eqref{eq:sparse-H}, i.e.,
\begin{gather*}
    H_{kk}=\Psi_k(A_k A_k^\top + B_k B_k^\top)\Psi_k=\Psi_k^2, \\
    H_{k-1,k}=-\Psi_k B_k A_{k-1}^\top \Psi_{k-1}.
\end{gather*}
Moreover, we can construct the multiplier $\Lambda_k=\frac{1}{2}\Psi_k^2$ and the weight matrix
\begin{equation}\label{eq:W-lbdn}
    W_k=-\Lambda_k^{-1}H_{k-1,k}=2\Psi_k^{-1}B_k A_{k-1}^\top \Psi_{k-1}
\end{equation}
with $k=0,\ldots,L$, where $A_{-1}=I,\Psi_{-1}=\sqrt{\gamma/2}I$ and $\Psi_{L}=\sqrt{2/\gamma}I$ with $\gamma$ as the prescribed Lipschitz bound.

\begin{figure}[]
\vskip 0.2in
\begin{center}
\centerline{\includegraphics[width=\linewidth]{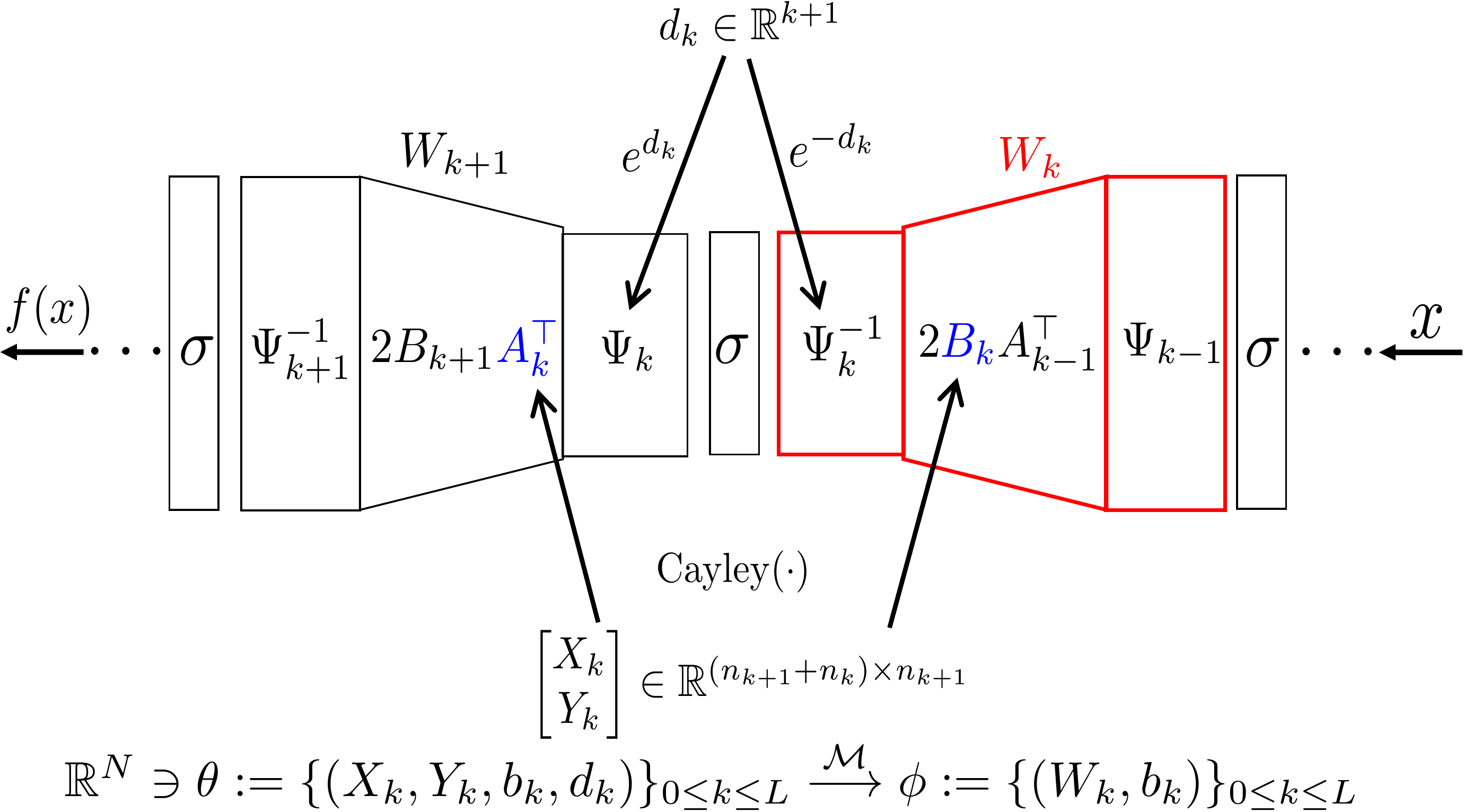}}
\caption{Direct parameterization for Lipschitz-bounded deep networks, i.e. $\mathrm{Lip}(f_{\phi})\leq \gamma$ with $\phi=\mathcal{M}(\theta)$ for all $\theta \in \R^N$. Note that free parameters are shared across neighbouring layers.}
\label{fig:lbdn}
\end{center}
\vskip -0.2in
\end{figure}

We summarize our model parameterization as follows. The free parameter $\theta$ consists of bias terms $b_k\in \R^{n_{k+1}}$ and 
\begin{gather*}
    d_j\in \R^{n_{j}}, \;
    X_k\in\R^{n_{k+1}\times n_{k+1}}, \;Y_k\in \R^{n_{k}\times n_{k+1}}
\end{gather*}
with $0\leq j< L$ and $0\leq k\leq L$. The weight $W_k$ is constructed  via \eqref{eq:PAB} and \eqref{eq:W-lbdn}. Notice that $W_k$ depends on parameters of index $k$ and $k-1$, i.e. there is an ``interlacing'' coupling between parameters and weights, see \cref{fig:lbdn}.

\subsection{Theoretical results}
The main theoretical results is that our parameterization is \emph{complete} (necessary and sufficient) for the set of DNNs satisfying the LMI constraint \eqref{eq:sparse-H} of \citep{fazlyab2019efficient}.
\begin{theorem}\label{thm:lbdn}
    The feedforward network \eqref{eq:DNN} satisfies the LMI condition \eqref{eq:sparse-H} if and only if its weight matrices $W_k$ can be parameterized via \eqref{eq:W-lbdn}.
\end{theorem}
The proof to this and all other theorems can be found in \cref{sec:proof}.

\paragraph{1-Lipschitz sandwich layer.} Here we show that the new parameterization can also be interpreted as a new layer type. We first introduce new hidden units $h_k=\sqrt{2}A_{k}^\top \Psi_k z_k$ for $k=0,\ldots L$ and then rewrite the proposed LBDN model as 
\begin{equation}\label{eq:lbdn}
    \begin{split}
        h_0=&\sqrt{\gamma} x \\
        h_{k+1}=&\sqrt{2}A_k^\top \Psi_k \sigma(\sqrt{2}\Psi_k^{-1}B_k h_k +b_k)\\
        y=&\sqrt{\gamma}B_{L}h_L+b_L.
    \end{split}
\end{equation}
The core component of the above model is a sandwich-structured layer of the form:
\begin{equation}\label{eq:sandwich-layer}
    \hout =\sqrt{2}A^\top \Psi \sigma\bigl(\sqrt{2}\Psi^{-1}B \hin +b\bigr)
\end{equation}
where $\hin \in \R^p, \hout \in \R^q$ are the layer input and output, respectively. Unlike the parameterization in \eqref{eq:W-lbdn}, consecutive layers in \eqref{eq:lbdn} do not have coupled free parameters, which allows for modular implementation. Another advantage is that such representation can reveal some fundamental insights on the roles of $\Psi,A$ and $B$.

\begin{theorem}\label{thm:layer}
    The layer \eqref{eq:sandwich-layer} with $\Psi,A,B$ constructed by \eqref{eq:PAB} is 1-Lipschitz.
\end{theorem}

To understand the role of $\Psi$, we look at a new activation layer which is obtained by placing $\Psi^{-1}\in \D_{++}^q$ and $\Psi$ before and after $\sigma$, i.e., $u=\Psi\sigma(\Psi^{-1}v+b)$. Here $\Psi$ can change the shape and shift the position of individual activation channels while keeping their slopes within $[0,1]$, allowing the optimizer to search over a rich set of activations.

For the roles of $A$ and $B$, we need to look at another special case of \eqref{eq:sandwich-layer} where $\sigma$ is the identity operator. Then, \eqref{eq:sandwich-layer} becomes a linear layer
\begin{equation}\label{eq:lip1-linear}
    \hout=2A^\top B\hin+\hat{b}.
\end{equation}
As a direct corollary of \cref{thm:layer}, the above linear layer is 1-Lipschitz, i.e., $\|2A^\top B\|\leq 1$. We show that such a parameterization is \emph{complete} for 1-Lipschitz linear layers.
\begin{proposition}\label{prop:linear-layer}
    A linear layer is 1-Lipschitz if and only if its weight $W$ satisfies $W=2A^\top B$ with $A,B$ given by \eqref{eq:PAB}.
\end{proposition}

\begin{algorithm}[tb]
   \caption{1-Lipschitz convolutional layer}
   \label{alg:conv}
\begin{algorithmic}[1]
   \REQUIRE $\hin\in \R^{p\times s\times s}$, $P\in \R^{(p+q)\times q\times s\times s}$, $d\in \R^{q}$
   \STATE \vspace*{0.2\baselineskip} $\tilde{h}_{\mathsf{in}}\leftarrow \FFT(\hin)$
   \STATE \vspace*{0.2\baselineskip} $\Psi\leftarrow \diag(e^d),\; \begin{bmatrix}
       \tilde A & \tilde B
   \end{bmatrix}^*\leftarrow\cayley(\FFT(P))$
   \STATE \vspace*{0.2\baselineskip} $\tilde{h}[:,i,j]\leftarrow \sqrt{2}\Psi^{-1}\tilde{B}[:,:,i,j]\tilde{h}_{\mathsf{in}}[:,i,j]$
   \STATE \vspace*{0.2\baselineskip} $\tilde{h}\leftarrow \FFT\bigl(\sigma(\FFT^{-1}(\tilde{h})+b)\bigr)$
   \STATE \vspace*{0.2\baselineskip} $\tilde{h}_{\mathsf{out}}[:,i,j]\leftarrow \sqrt{2}A[:,:,i,j]^*\Psi \tilde{h}[:,i,j]$
   \STATE \vspace*{0.2\baselineskip} $\hout\leftarrow \iFFT(\tilde{h}_{\mathsf{out}})$
\end{algorithmic}
\end{algorithm}

\paragraph{Convolution layer.} Our proposed 1-Lipschitz layer can also incorporate more structured linear operators such as \emph{circular convolution}. Thanks to the doubly-circular structure, we can perform efficient model parameterization in the Fourier domain. Roughly speaking, transposing or inverting a convolution is equivalent to apply the complex version of the same operation to its Fourier domain representation -- a batch of small complex matrices  \cite{trockman2021orthogonalizing}. \cref{alg:conv} shows the forward computation of 1-Lipschitz convolutional layers, see \cref{sec:conv} for more detailed explanations. In line 1 and 6, we use the (inverse) FFT on the input/output tensor. In line 2, we perform the Cayley transformation of convolutions in the Fourier domain, which involves $s\times (\lfloor s/2\rfloor+1)$ parallel complex matrix inverse of size $q\times q$ where $q,s$ are the number of hidden channels and input image size, respectively. In line 3-5, all operations related to the $(i,j)^{\mathrm{th}}$ term can be done in parallel.

%%%%%%%%%%%%%%%%%%%%%%%%%%%%%%%%%%%%%%%%%%%%%%%%%%%%%%%%%%%%%%%%%%%%%%%%%%%%%%%
\section{Comparisons to Related Prior Work}

In this section we give an overview of related prior work and provide some theoretical comparison to the proposed approach.

\subsection{SDP-based Lipschitz training}
Since the SDP-based bounds of \citep{fazlyab2019efficient} appeared, several papers have proposed methods to allow training of Lipschitz models. In \citep{pauli2021training,revay2020convex}, training was done by constrained optimization techniques (projections and barrier function, respectively). However, those approaches have the computational bottleneck for relatively-small ($\sim$1000 neurons) networks. \cite{xue2022chordal} decomposed the large SDP for Lipschitz bound estimation into many small SDPs via chordal decomposition.

Direct parameterization of the SDP-based Lipschitz condition was introduced in \cite{revay2020lipschitz} for equilibrium networks -- a more general architecture than the feedforward networks. The basic idea was related to the method of \cite{burer2003nonlinear} for semi-definite programming, in which a positive semi-definite matrix is parameterized by square-root factors. In \cite{revay2021recurrent}, it was further extended to recurrent (dynamic) equilibrium networks. \cite{pauli2022lipschitz,pauli2023lipschitz} applied this method to Lipschitz-bound 1D convolutional networks. However, those approaches do not scale to large DNNs. In this work, we explore the sparse structure of SDP condition for DNNs, leading to a scalable direct parameterization. A recent work \cite{araujo2023unified} also developed a scalable parameterization for training residual networks. But its Lipschitz condition only considers individual layer, which is often a relatively small SDP with dense structure.

\subsection{1-Lipschitz neural networks}

Many existing works have focused on the construction of provable $1$-Lipschitz neural networks. Most are bottom-up approaches, i.e., devise 1-Lipschitz layers first and then connect them in a feedforward way. One approach is to build 1-Lipschitz linear layer $ z=Wx$ with $\|W\|\leq 1$ since most existing activation layers are 1-Lipschitz (possibly after rescaling). The Lipschitz bound is quite loose due to the decoupling between linear layer and nonlinear activation. Another direction is to construct $1$-Lipschitz layer which directly involves activation function. 

\paragraph{$1$-Lipschitz linear layers.}  Early works \cite{miyatospectral,farniageneralizable} involve layer normalization via spectral norm:
\[
W=V/\|V\|
\]
with $V$ as free parameter. Some recent works construct gradient preserved linear layer by constraining $W$ to be orthogonal during training, e.g., block convolution orthogonal parameterization \cite{li2019preventing}, orthogonal matrix parameterization via Cayley transformation \cite{trockman2021orthogonalizing,yu2022constructing}, matrix exponential \cite{singla2021skew}
\[
W=\exp(V-V^\top)
\]
and inverse square root \cite{xu2022lot}
\[
W=(VV^\top)^{-1} V.
\]
Almost Orthogonal Layer (AOL) \cite{prach2022almost} can reduce the computational cost by using the inverse of a diagonal matrix, i.e., \[
W=V\diag\bigl(\sum\nolimits_j |V^\top V|_{ij}\bigr)^{-1/2}.
\]
Empirical study reveals that $W$ is almost orthogonal after training. For these approaches, the overall network Lipschitz bound is then obtained via a spectral norm bound: 
\begin{equation*}
    \|\mathrm{Lip}(f)\|\leq \prod_{k=0}^{L}\|W_k\|\leq 1.
\end{equation*}

Compared to 1-Lipschitz linear layers, our approach has two advantages. First, a special case of our sandwich layer \eqref{eq:lip1-linear} contains all 1-Lipschitz linear layers (see \cref{prop:linear-layer}). Second, our model parameterization allows for the spectral norm bounds of individual layers to be greater than one, and their product to also be greater than one, while the network still satisfies a Lipschitz bound of 1, see the example in \cref{fig:bound} as well as the explanation in \cref{sec:norm-bound}. 

\paragraph{1-Lipschitz nonlinear layers.} Since spectral-norm bounds can be quite loose, a number of recent papers have constructed Lipschitz-bounded nonlinear layers. In \cite{meunier2022dynamical}, a new $1$-Lipschitz residual-type layer $z=x+\mathcal{F}(x)$ with $\mathcal{F}(x)=-2/\|W\|^2 W\sigma(W^\top x+b)$, is derived from dynamical systems called convex potential flows. Recently, \cite{araujo2023unified} considered a more general layer:
\begin{equation}\label{eq:sll-layer}
    h_s(x)=Hx+G\sigma(W^\top x+b),
\end{equation} 
and provides an extension to the SDP condition in \cite{fazlyab2019efficient} as follows 
\begin{equation}\label{eq:sdp-sll}
\begin{bmatrix}
   \gamma I-H^\top H & -H^\top G- W\Lambda \\ -G^\top H -\Lambda W^\top & 2\Lambda - \frac{1}{\gamma}G^\top G 
\end{bmatrix}\succeq 0.
\end{equation}
For the special case with $\gamma=1$ and $H=I$,  a direct parameterization of \eqref{eq:sdp-sll} is $G=-2W T^{-1}$, where $W$ is a free variable and $T\in \D_{++}$ satisfies $T\succeq W^\top W$. The corresponding $h_s(x)$ is called SDP-based Lipschitz Layer (SLL). Similar to the SLL approach, our proposed sandwich layer \eqref{eq:sandwich-layer} can also be understood as an analytical solution to \eqref{eq:sdp-sll} but with a different case with $H=0$ and arbitrary $\gamma$. 

Moreover, \cref{thm:lbdn} shows that by composing many 1-Lipschitz sandwich layers and then adding a scaling factor $\sqrt{\gamma}$ into the first and last layers, we can construct all the DNNs satisfying the (structured, network-scale) SDP in \cite{fazlyab2019efficient} for any Lipschitz bound $\gamma$. When an SLL layer with $\gamma>1$ is desired, similarly one can compose an 1-Lipschitz SLL layer with $\gamma$, i.e.
\begin{equation}\label{eq:scaled-sll}
    h(x)=\gamma^{p} h_s(\gamma^q x)
\end{equation}
with $p+q=1$ and $p,q\ge 0$.  However, such a parameterization is incomplete as the example below gives a residual layer which satisfies \eqref{eq:sdp-sll}, but cannot be constructed via \eqref{eq:scaled-sll}.   
\begin{example}
Consider the following following residual layer, which has a Lipschitz bound of $1.001$:
\[
h(x)=x+\begin{bmatrix}
    1 & 0 \\ 0 & 0.001
\end{bmatrix}\sigma\left(
\begin{bmatrix}
    0 & 0 \\ 0 & 1
\end{bmatrix}x+b
\right).
\]
It can be verified that \eqref{eq:sdp-sll} is satisfied with $\gamma=1.001$ and $\Lambda=\diag(\lambda_1,\lambda_2)$ chosen such that $\lambda_1>(\gamma-1/2)/(\gamma^2-\gamma)$ and $\lambda_2=\gamma^2-\gamma$. However, it cannot be written as \eqref{eq:scaled-sll} because there does not exist a positive diagonal matrix $T$ such that $G=-2W T^{-1}$, since the upper-left element is zero in $W$ and one in $G$.
\end{example}

%%%%%%%%%%%%%%%%%%%%%%%%%%%%%%%%%%%%%%%%%%%%%%%%%%%%%%%%%%%%%%%%%%%%%%%%%%%%%%%
\section{Experiments}

Our experiments have two goals: First, to illustrate that our model parameterization can provide a tight Lipschitz bounds via a simple curve-fitting tasks. Second, to examine the performance and scalability of the proposed method on robust image classification tasks. Model architectures and training details can be found in \cref{sec:training-details}. Pytorch code is available at \url{https://github.com/acfr/LBDN}, and a partial implementation of the method is included in the Julia toolbox {\tt RobustNeuralNetworks.jl}, \url{https://acfr.github.io/RobustNeuralNetworks.jl}

\subsection{Toy example}\label{sec:toy}

\begin{table}[t]
\caption{The table presents the tightness of Lipschitz bound of several concurrent parameterization and our approach on a toy example. The bound tightness is measured by $\underline{\gamma}/\gamma$ (\%), where $\underline{\gamma}$ and $\gamma$ are the empirical lower bound and certified upper bound.}
\label{tab:TightLip}
\vskip 0.15in
\begin{center}
\begin{small}
\begin{sc}
\begin{tabular}{lrcccc}
    \toprule
    \multicolumn{1}{c}{\multirow{2}[4]{*}{\textbf{Models}}} &   & \multicolumn{3}{c}{\textbf{Lip. tightness ($\gamma$)}}  \\
\cmidrule{3-5}      &   & 1 & 5 & 10   \\
    \midrule
    \textbf{AOL}        & & 77.2 & 45.2 & 47.9 \\ 
    \textbf{Orthogonal} & & 74.1 & 72.8 & 64.5 \\ 
    \textbf{SLL}        & & 99.9 & 90.5 & 67.9 \\ 
    \textbf{Sandwich}   & & 99.9 & 99.3 & 94.0 \\ 
    \bottomrule
    \end{tabular}%
\end{sc}
\end{small}
\end{center}
\vskip -0.1in
\end{table}

\begin{figure}[t]
\vskip 0.2in
\begin{center}
\def\figwidth{0.81\columnwidth}
\def\figheight{0.18\textheight}
\input{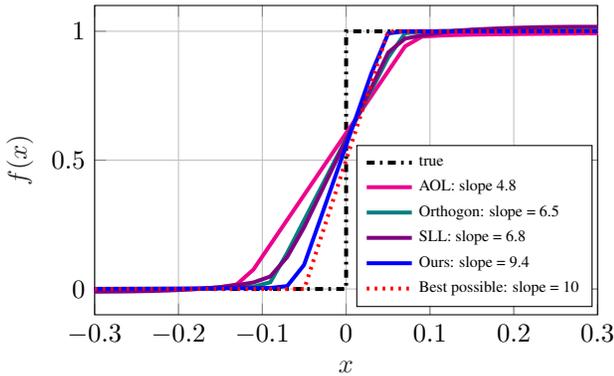}
\caption{Fitting a square wave using models with Lipschitz bound of 10. Compared to AOL, orthogonal and SLL layers, our model is the closest to the best possible solution -- a piecewise linear function with slope of 10 at $x=0$.} 
\label{fig:square-wave}
\end{center}
\vskip -0.2in
\end{figure}

We illustrate the quality of Lipschitz bounds of by fitting the following square wave: 
\[
f(x)=
    \begin{cases}
        0, & x\in [-1,0)\cup [1,2], \\
        1, & x\in [-2,-1) \cup [0,1).
    \end{cases}
\]

Note that the true function has no global Lipschitz bound due to the points of discontinuity. Thus a function approximator will naturally try to find models with large local Lipschitz constant near the discontinuity. If a Lipschitz bound is imposed this is a useful test of its accuracy, which wee evaluate  using $\underline{\gamma}/\gamma$ where $\underline{\gamma}$ is an empirical lower Lipschitz bound and $\gamma$ is the imposed upper bound, being 1, 5, and 10 in the cases we tested. In Table \ref{tab:TightLip} we see that  our approach achieves a much tighter Lipschitz bounds than AOL and orthogonal layers. The SLL model has similar tightness when $\gamma=1$ but its bound becomes more loose as $\gamma$ increases compared to our model, e.g. 67.9\% v.s. 94.0\% for $\gamma=10$. We also plot the fitting results for $\gamma=10$ in \cref{fig:square-wave}. Our model is  close to the best possible solution: a piecewise linear function with slope 10 at the discontinuities.

\begin{figure}[th]
\vskip 0.2in
\begin{center}
\def\figwidth{0.37\columnwidth}
\def\figheight{0.15\textheight}
% \documentclass[11pt]{standalone}

% \usepackage{amsmath}
% \usepackage{graphicx}
% \usepackage{pgfplots}
% \pgfplotsset{compat=newest}
% \pgfplotsset{scaled y ticks=false}
% \usepgfplotslibrary{groupplots}
% \usepgfplotslibrary{dateplot}
% \pgfplotsset{compat=1.11,
% 	every axis plot/.append style={line width=1pt},
% 	legend style={font=\tiny}
% }

% \usepackage{tikz}
% \usetikzlibrary{plotmarks}

% \begin{document}

% \def\figwidth{3.6cm}
% \def\figheight{4.2cm}

\begin{tikzpicture}

\begin{axis}[%
height=\figheight,
width=\figwidth,
name=plot121,
scale only axis,
ymin=1,
ymax=11,
xtick={75, 80, 90, 100},
xticklabels={{\scriptsize 75}, {\scriptsize 80}, {\scriptsize 90}, {\scriptsize 100}},
ytick={1,2,3,4,5,6,7,8,9,10,11},
yticklabels={{\scriptsize Input layer},{\scriptsize Hidden layer 1},{\scriptsize Hidden layer 2},{\scriptsize Hidden layer 3},{\scriptsize Hidden layer 4},{\scriptsize Hidden layer 5},{\scriptsize Hidden layer 6},{\scriptsize Hidden layer 7},{\scriptsize Hidden layer 8},{\scriptsize Output layer},{\scriptsize Full network}},
yticklabel style={rotate=0},
% xmode=log,
xmin=70,
xmax=103,
xminorticks=true,
xlabel style={font=\color{white!15!black}},
xlabel={\scriptsize $\underline{\gamma}/\gamma$ (\%)},
axis background/.style={fill=white},
xmajorgrids,
ymajorgrids,
legend style={legend cell align=left, align=left, draw=white!15!black},
legend style={nodes={scale=1.4, transform shape}, draw=black},
legend pos=south west
]
\addplot [color=blue, mark=x, only marks, mark size=3, mark options={solid, blue}]
  table[row sep=crcr]{%
% Layer Lip. Sandwich
99.9900 1\\
99.3500 2\\
99.3700 3\\
99.1900 4\\
99.1900 5\\
98.9100 6\\
99.5900 7\\
99.3300 8\\
99.5300 9\\
99.9200 10\\
99.9600 11\\
};
\addlegendentry{Sandwich}

\addplot [color=teal, mark=o, only marks, mark size=3, mark options={solid, teal}]
  table[row sep=crcr]{%
% Layer Lip. Orthogonal
98.9100 1\\
99.8600 2\\
99.8700 3\\
99.8800 4\\
99.9000 5\\
99.9100 6\\
99.9100 7\\
99.9000 8\\
99.9000 9\\
99.9000 10\\
74.1100 11\\
};
\addlegendentry{Orthogonal}

\addplot [color=magenta, mark=+, only marks, mark size=3, mark options={solid, magenta}]
  table[row sep=crcr]{%
% Layer Lip. AOL
100.0000 1\\
99.8000 2\\
99.7400 3\\
99.8200 4\\
99.7700 5\\
99.7800 6\\
99.7100 7\\
99.7700 8\\
99.6700 9\\
99.9300 10\\
77.1900 11\\
};
\addlegendentry{AOL}

\end{axis}

\begin{axis}[%
width=\figwidth,
height=\figheight,
name=plot122,
at=(plot121.right of south east), 
anchor=left of south west,
scale only axis,
% xmode=log,
xmin=0.7,
xmax=2.5,
xminorticks=true,
xtick={1, 2, 2.5},
xticklabels={{\scriptsize $1$}, {\scriptsize $2$}, {\scriptsize $2.5$}},
xlabel style={font=\color{white!15!black}},
xlabel={\scriptsize $\|W\|$},
ymin=1,
ymax=11,
ytick={1,2,3,4,5,6,7,8,9,10,11},
yticklabels={{ }, { }, { }, { }, { }, { }, { }, { }, { }, { }, { }},
axis background/.style={fill=white},
xmajorgrids,
% xminorgrids,
ymajorgrids,
% legend style={legend cell align=left, align=left, draw=white!15!black},
% legend pos=south east
]
\addplot [color=blue, only marks, mark=x, mark size = 3, mark options={solid, blue}]
  table[row sep=crcr]{%
% Layer Norm. Sandwich
0.9615 1\\
1.9136 2\\
1.7568 3\\
1.9932 4\\
2.0420 5\\
1.4883 6\\
1.6785 7\\
1.7198 8\\
1.5306 9\\
0.7623 10\\
};
% \addlegendentry{Sandwich}

\addplot [color=teal, only marks, mark=o, mark size = 3, mark options={solid, teal}]
  table[row sep=crcr]{%
1.0000 1\\
1.0000 2\\
1.0000 3\\
1.0000 4\\
1.0000 5\\
1.0000 6\\
1.0000 7\\
1.0000 8\\
1.0000 9\\
1.0000 10\\
};
% \addlegendentry{Orthogon}

\addplot [color=magenta, only marks, mark=+, mark size = 3, mark options={solid, magenta}]
  table[row sep=crcr]{%
1.0000 1\\
0.9997 2\\
0.9999 3\\
0.9999 4\\
0.9999 5\\
1.0000 6\\
1.0000 7\\
1.0000 8\\
1.0000 9\\
1.0000 10\\
};

\end{axis}
\end{tikzpicture}%

% \end{document}
\caption{{\bf Left:} empirical Lipschitz bound for curve fitting of a square wave. The lower bound $\underline{\gamma}$ is obtained using PGD-like method. We observed tight layer Lipschitz bound for AOL, orthogonal and sandwich layers ($\geq 99.1\%$). However, the propose sandwich layer has a much tighter Lipschitz bound for the entire network. {\bf Right:} the spectral norm of weight matrices. Our approach admits weight matrices with spectral norm larger than 1. The layerwise product $\prod_{k=0}^{L}\|W_k\|$ is about 65.9, which is much larger than that of AOL and orthogonal layers.}
\label{fig:bound}
\end{center}
\vskip -0.2in
\end{figure}
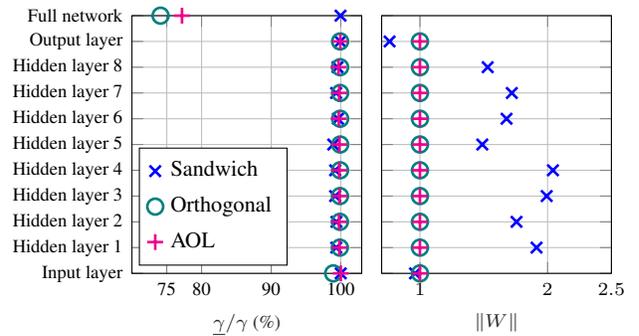

In \cref{fig:bound} we break down the Lipschitz bounds and spectral norms over layers. Note that the SLL model is not included here as its Lipschitz bound is not related to the spectral norms. It can be seen that all individual layers have quite tight Lipschitz bounds on a per-layer basis of around $99\%$.  However, for the complete network the sandwich layer achieves a much tighter bound of 99.9\% vs 74.1\% (orthogonal) and 77.2\% (AOL). This illustrates the benefits of taking into account coupling between neighborhood layers, thus allowing individual layers to have spectral norm greater than 1. We note that, for the sandwich model, the layer-wise product of spectral norms reaches 65.9, illustrating how poor this commonly-used bound is compared to our bound.

\begin{figure*}[ht]
\vskip 0.2in
\begin{center}
\def\figwidth{0.43\columnwidth}
\def\figheight{0.15\textheight}
\input{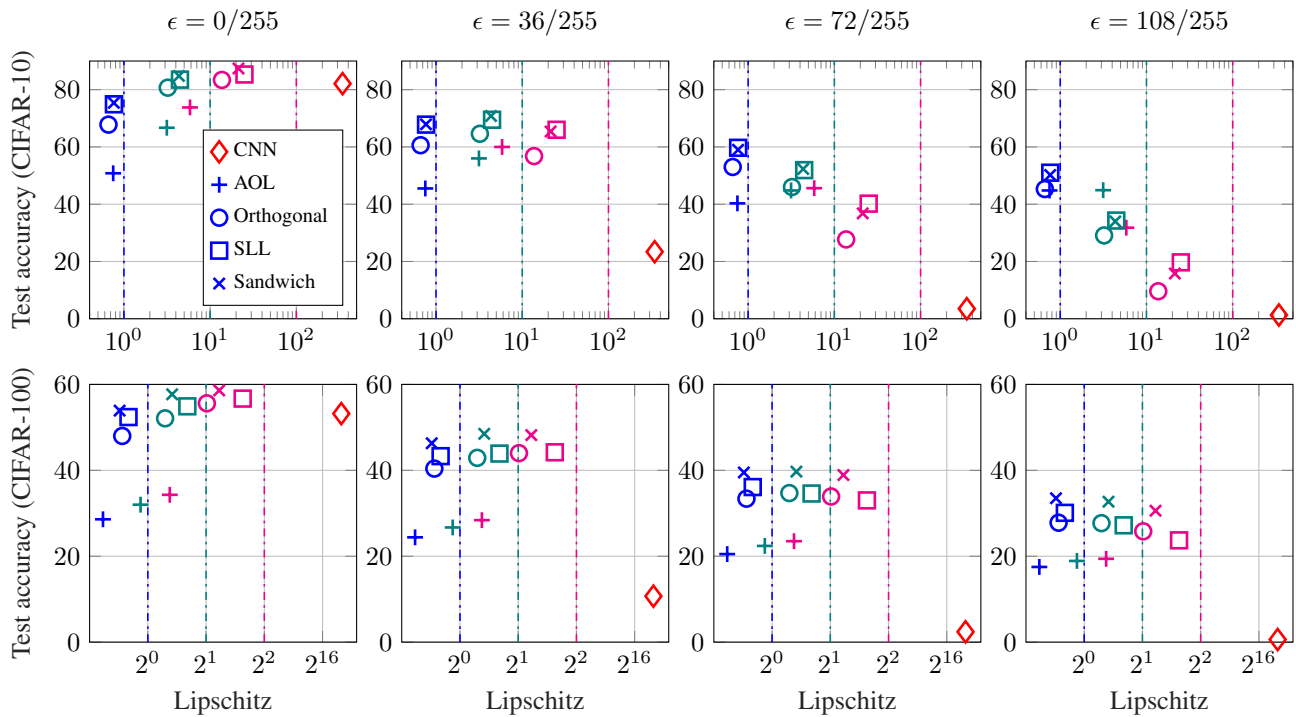}
\caption{Robust test accuracy under different $\ell^2$-adversarial attack sizes \emph{versus} empirical Lipschitz bound on CIFAR-10. Colours: blue ($\gamma=1$), teal ($\gamma=10$), magenta ($\gamma=100$), red (vanilla CNN). Vertical lines are the certified Lipschitz bounds. The empirical robustness is measured with \emph{AutoAttack} \cite{croce2020reliable}. For CIFAR-10, the SLL layer slightly outperforms the proposed sandwich layer but its model is much larger (41M versus 3M). But CIFAR-100, our model has about 4\% improvement dispite its relatively small model size compared to the SLL model (i.e. 48M versus 118M).
}
\label{fig:cifar10}
\end{center}
\vskip -0.2in
\end{figure*}

\subsection{Robust Image classification}
We conducted a set of empirical robustness experiments on CIFAR-10/100 and Tiny-Imagenet datasets, comparing our Sandwich layer to the previous parameterizations AOL, orthogonal and SLL layers. We use the same architecture in \cite{trockman2021orthogonalizing} for AOL, orthogonal and sandwich models with \emph{small, medium} and \emph{large} sizes. Since SLL is a residual layer, we use the architectures proposed by \cite{araujo2023unified}, with model sizes  much larger than those of the non-residual networks. Input data is normalized before feeding into Lipschitz bounded model. The Lipschitz bound for the composited model is fixed during the training. We use \emph{AutoAttack} \cite{croce2020reliable} to measure the empirical robustness. 

We also compare the certified robustness results of the proposed parameterization with the recently proposed SLL model  \cite{araujo2023unified}. We removed the data normalization layer and add a Last Layer Normalization (LLN), proposed by \cite{singlaimproved}. While the Lipschitz constant of the composite model may exceed the bound,  it has been observed in \cite{singlaimproved,araujo2023unified} that LLN can improved the certified accuracy when the number of classes becomes large.

\paragraph{Effect of Lipschitz bounds.} We trained Lipschitz-bounded models on CIFAR-10/100 datasets with three different certified bounds ($\gamma=1,10,100$ for CIFAR-10 and $\gamma=1,2,4$ for CIFAR-100). We also trained a vanilla CNN model without any Lipschitz regularization as a baseline. In \cref{fig:cifar10} we plot both the clean accuracy ($\epsilon=0$) and robust test accuracy under different $\ell^2$-adversarial attack sizes ($\epsilon=36/255, 72/255, 108/255$). The sandwich layer had higher test accuracy than the AOL and orthogonal layer in all cases, illustrating the improved flexibility. On CIFAR-10 our model is slightly outperformed by the the SLL model, although the model size of the latter is much larger (3M vs 41M parameters). On CIFAR-100, our model outperforms SLL by about 4\%  despite a much smaller model size (48M vs 118M). 

It can be seen that with an appropriate Lipschitz bound, all models except AOL had improved nominal test accuracy (i.e. $\epsilon=0$) compared to a vanilla CNN. This performance deteriorates if the Lipschitz bound is chosen to be too small. On the other hand, when the perturbation size is large (e.g. $\epsilon = 72/255$ or $108/255$), the smallest Lipschitz bounds yielded the best performance (except for the AOL). Furthermore, with these larger attack sizes, the performance improvement compared to vanilla CNN is very significant, e.g. close to 60\% on CIFAR10 with $\epsilon=72/255$.

\begin{table*}[t]
\centering
\caption{This table presents the clean, empirical robust accuracy as well as the number of parameters and training time of several concurrent work and our sandwich model on CIFAR-100 and Tiny-ImageNet datasets. Input data is normalized and no last layer normalization is implemented. The Lipschitz bounds for CIFAR-100 and Tiny-ImageNet are 2 and 1, respectively. The empirical robustness is measured with \emph{AutoAttack} \cite{croce2020reliable}. All results are averaged of 3 experiments.}\label{tab:emp-robust}
\vskip 0.15in
\begin{center}
\begin{small}
\begin{sc}
    \begin{tabular}{llrrrrrr}
    \toprule
    \multicolumn{1}{c}{\multirow{2}[2]{*}{\textbf{Datasets}}} & \multicolumn{1}{c}{\multirow{2}[2]{*}{\textbf{Models}}} & \multicolumn{1}{c}{\multirow{2}[2]{*}{\textbf{\makecell{\textbf{Clean} \\ \textbf{Acc.}}}}} & \multicolumn{3}{c}{\boldmath{}\textbf{AutoAttack ($\varepsilon$)}\unboldmath{}} & \multicolumn{1}{c}{\multirow{2}[2]{*}{\textbf{\makecell{\textbf{Number of} \\ \textbf{Parameters}}}}} & \multicolumn{1}{c}{\multirow{2}[2]{*}{\textbf{\makecell{\textbf{Time by} \\ \textbf{Epoch}}}}} \\
   \cmidrule{4-6}
     & &   & $\frac{36}{255}$ & $\frac{72}{255}$ & $\frac{108}{255}$  \\
    \midrule
    \multirow{11}[4]{*}{CIFAR100}
      & \textbf{AOL Small} & 30.4 & 25.1 & 21.1 & 17.6 & 3M & 18s \\ 
    & \textbf{AOL Medium} & 31.1 & 25.9 & 21.7 & 18.2 & 12M & 21s \\ 
    & \textbf{AOL Large} & 31.6 & 26.5 & 22.2 & 18.7 & 48M & 73s \\ 
    & \textbf{Orthogonal Small} & 48.7 & 38.6 & 30.6 & 24.0 & 3M & 20s \\ 
    & \textbf{Orthogonal Medium} & 51.1 & 41.4 & 33.0 & 26.4 & 12M & 22s \\ 
    & \textbf{Orthogonal Large} & 52.2 & 42.5 & 34.3 & 27.4 & 48M & 55s \\ 
    & \textbf{SLL Small} & 52.9 & 41.9 & 32.9 & 25.5 & 41M & 29s \\ 
    & \textbf{SLL Medium} & 53.8 & 43.1 & 33.9 & 26.6 & 78M & 52s \\ 
    & \textbf{SLL Large} & 54.8 & 44.0 & 34.9 & 27.6 & 118M & 121s \\ 
    \cmidrule{2-8}
      & \textbf{Sandwich Small} & 54.2 & 44.3 & 35.5 & 28.4 & 3M & 19s \\ 
    & \textbf{Sandwich Medium} & 56.5 & 47.1 & 38.6 & 31.5 & 12M & 23s \\ 
    & \textbf{Sandwich Large} & \textbf{57.5} & \textbf{48.5} & \textbf{40.2} & \textbf{32.9} & 48M & 78s \\ 
    \midrule
    \multirow{7}[4]{*}{TinyImageNet}
      & \textbf{AOL Small} & 17.4 & 15.1 & 13.1 & 11.3 & 11M & 62s \\ 
    & \textbf{AOL Medium} & 16.8 & 14.6 & 12.7 & 11.0 & 43M & 270s \\ 
    & \textbf{Orthogonal Small} & 29.7 & 24.4 & 20.1 & 16.4 & 11M & 57s \\ 
    & \textbf{Orthogonal Medium} & 30.9 & 26.0 & 21.5 & 17.7 & 43M & 89s \\ 
    & \textbf{SLL Small} & 29.3 & 23.5 & 18.6 & 14.7 & 165M & 203s \\ 
    & \textbf{SLL Medium} & 30.3 & 24.6 & 19.8 & 15.7 & 314M & 363s \\ 
    \cmidrule{2-8}
      & \textbf{Sandwich Small} & 34.7 & 29.3 & 24.6 & 20.5 & 10M & 60s \\ 
    & \textbf{Sandwich Medium} & \textbf{35.0} & \textbf{29.9} & \textbf{25.3} & \textbf{21.4} & 37M & 139s \\
    \bottomrule
    \end{tabular}%
\end{sc}
\end{small}
\end{center}
\vskip -0.1in
\end{table*}%

\begin{figure}[h]
    \centering
    \def\figwidth{0.8\columnwidth}
    \def\figheight{0.15\textheight}
    \input{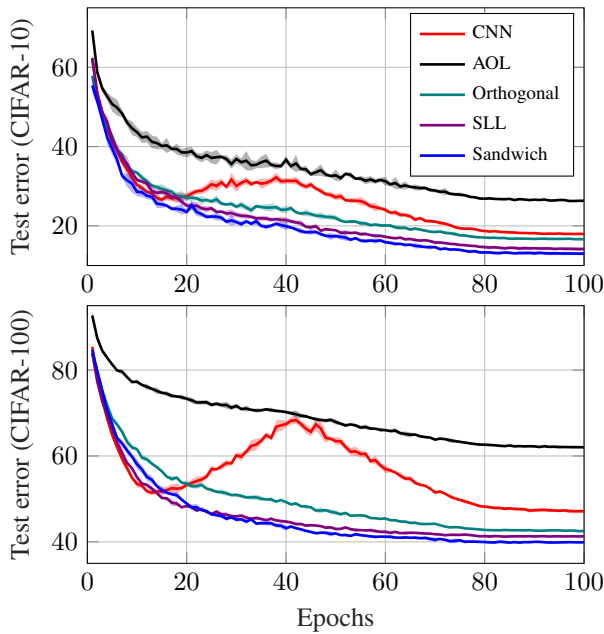}
    \caption{Learning curves (obtained from 5 experiments). We use $\gamma=100$ and $10$ for CIFAR-10/100, respectively. The ``double-descent'' phenomenon is avoided with the $\gamma$-Lipschitz models.}
    \label{fig:cifar10-train}
\end{figure}

In \cref{fig:cifar10-train} we plot the training curves (test-error vs epoch) for the Lipschitz-bounded and vanilla CNN models. We observe that the sandwich model surpasses the final error of CNN in less than half as many epochs. An interesting observation from \cref{fig:cifar10-train} is that the CNN model seems to exhibit the epoch-wide double descent phenomenon (see, e.g., \cite{nakkiran2021deep}), whereas none of the Lipschitz bounded models do, they simply improve test error monotonically with epochs. Weight regularization has been suggested as a mitigating factor for other forms of double descent \cite{nakkiranoptimal}, however we are not aware of this specific phenomenon having been observed before.

\begin{figure}[h]
\vskip 0.2in
\begin{center}
\def\figwidth{0.8\columnwidth}
\def\figheight{0.15\textheight}
% \documentclass[11pt]{standalone}

% \usepackage{amsmath}
% \usepackage{graphicx}
% \usepackage{pgfplots}
% \pgfplotsset{compat=newest}
% \pgfplotsset{scaled y ticks=false}
% \usepgfplotslibrary{groupplots}
% \usepgfplotslibrary{dateplot}
% \pgfplotsset{compat=1.11,
% 	every axis plot/.append style={line width=1pt},
% 	legend style={font=\tiny}
% }

% \usepackage{tikz}
% \usetikzlibrary{plotmarks}

% \begin{document}

% \def\figwidth{5cm}
% \def\figheight{4cm}

\begin{tikzpicture}

\begin{axis}[
width=\figwidth,
height=\figheight,
name=plot121,
at={(0,1)},
scale only axis,
%legend pos=south east,
legend style={nodes={scale=1.2, transform shape}, draw=black},
legend style={at={(0.98,0.05)},anchor=south east},
legend cell align=left,
xmin=0,
xmax=120,
xminorticks=true,
ymin=20,
ymax=60,
xlabel style={font=\color{white!15!black}},
ylabel style={font=\color{white!15!black}},
ylabel={Accuracy (CIFAR-100)},
axis background/.style={fill=white},
xmajorgrids,
ymajorgrids
]

\addplot[only marks, mark=+, mark options={}, mark size=3, draw=blue] table[row sep=crcr]{%
x	y\\
3	30.4\\
};

\addplot[only marks, mark=o, mark options={}, mark size=3, draw=blue] table[row sep=crcr]{%
x	y\\
3	48.7\\
};

\addplot[only marks, mark=square, mark options={}, mark size=3, draw=blue] table[row sep=crcr]{%
x	y\\
41	52.9\\
};

\addplot[only marks, mark=x, mark options={},mark size=3, draw=blue] table[row sep=crcr]{%
x	y\\
3	54.2\\
};

\addplot[only marks, mark=+, mark options={}, mark size=3, draw=teal, forget plot] table[row sep=crcr]{%
x	y\\
12	31.1\\
};

\addplot[only marks, mark=o, mark options={}, mark size=3, draw=teal, forget plot] table[row sep=crcr]{%
x	y\\
12	51.1\\
};

\addplot[only marks, mark=square, mark options={}, mark size=3, draw=teal, forget plot] table[row sep=crcr]{%
x	y\\
78	53.8\\
};

\addplot[only marks, mark=x, mark options={}, mark size=3, draw=teal, forget plot] table[row sep=crcr]{%
x	y\\
12	56.5\\
};

\addplot[only marks, mark=+, mark options={}, mark size=3, draw=magenta, forget plot] table[row sep=crcr]{%
x	y\\
48	31.6\\
};

\addplot[only marks, mark=o, mark options={}, mark size=3, draw=magenta, forget plot] table[row sep=crcr]{%
x	y\\
48	52.2\\
};

\addplot[only marks, mark=square, mark options={}, mark size=3, draw=magenta, forget plot] table[row sep=crcr]{%
x	y\\
118	54.8\\
};

\addplot[only marks, mark=x, mark options={}, mark size=3, draw=magenta, forget plot] table[row sep=crcr]{%
x	y\\
48	57.5\\
};
\legend{AOL,Orthogonal,SLL,Sandwich}
\end{axis}

\begin{axis}[%
width=\figwidth,
height=\figheight,
name=plot122,
at=(plot121.below south west), 
anchor=above north west,
% at=(plot121.right of south east), 
% anchor=left of south west,
scale only axis,
%legend pos=south east,
legend style={nodes={scale=0.8, transform shape}, draw=black},
legend style={at={(0.93,0.05)},anchor=south east},
legend cell align=left,
xmin=0,
xmax=350,
xminorticks=true,
xlabel style={font=\color{white!15!black}},
xlabel={Number of parameters (M)},
ymin=10,
ymax=40,
ylabel style={font=\color{white!15!black}},
ylabel={Accuracy (Tiny-imagenet)},
axis background/.style={fill=white},
xmajorgrids,
ymajorgrids
]
\addplot[only marks, mark=+, mark options={}, mark size=3, draw=blue, forget plot] table[row sep=crcr]{%
x	y\\
11	17.4\\
};
\addplot[only marks, mark=+, mark options={}, mark size=3, draw=teal, forget plot] table[row sep=crcr]{%
x	y\\
43	16.8\\
};

\addplot[only marks, mark=o, mark options={}, mark size=3, draw=blue, forget plot] table[row sep=crcr]{%
x	y\\
11	29.7\\
};
\addplot[only marks, mark=o, mark options={}, mark size=3, draw=teal, forget plot] table[row sep=crcr]{%
x	y\\
43	30.9\\
};
\addplot[only marks, mark=square, mark options={}, mark size=3, draw=blue, forget plot] table[row sep=crcr]{%
x	y\\
165	29.3\\
};
\addplot[only marks, mark=square, mark options={}, mark size=3, draw=teal, forget plot] table[row sep=crcr]{%
x	y\\
314	30.3\\
};

\addplot[only marks, mark=x, mark options={}, mark size=3, draw=blue, forget plot] table[row sep=crcr]{%
x	y\\
10	34.7\\
};
\addplot[only marks, mark=x, mark options={}, mark size=3, draw=teal, forget plot] table[row sep=crcr]{%
x	y\\
37	35.0\\
};

\end{axis}

\end{tikzpicture}%

% \end{document}
\caption{Test accuracy \emph{versus} model size on CIFAR-100 and Tiny-Imagenet. Colours: blue (small), teal (medium), magenta (large). For CIFAR-100, our small sandwich model (3M) has the similar performance as the large SLL model (118M). For Tiny-Imagenet, our small sandwich model (10M) has about 4.5\% improvement in test accuracy compared to the medium SLL model (314M).} 
\label{fig:cifar100}
\end{center}
\vskip -0.2in
\end{figure}
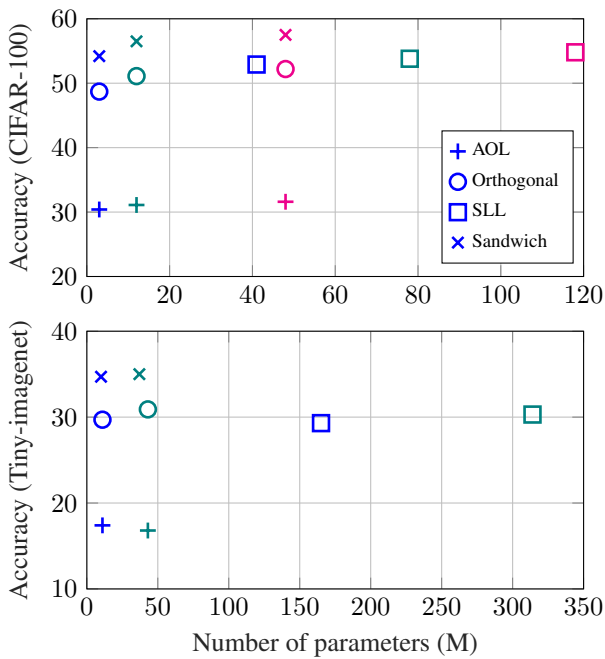

\paragraph{Empirical robustness on CIFAR-100 and Tiny-Imagenet.}
We ran empirical robustness tests on larger datasets  (CIFAR-100 and Tiny-Imagenet). We trained models with Lipschitz bounds of  $\{0.5,1,\ldots,16 \}$ and presented the one with best robust accuracy for $\epsilon=36/255$. The results along with total number of parameters (NP) and training time per epoch (TpE) are collected in \cref{tab:emp-robust}. We also plot the test accuracy versus model size in \cref{fig:cifar100}.

We observe that our proposed Sandwich layer achieves uniformly the best results (around 5\% improvement) on both CIFAR-100 and Tiny-Imagenet for all model sizes, in terms of both clean accuracy and robust accuracy with all perturbation sizes. Furthermore, our Sandwich model can achieve superior results with much smaller models and faster training than SLL. On CIFAR-100, comparing our Sandwich-medium vs SLL-large we see that ours gives superior clean and robust accuracy despite having only 12M parameters vs 118M, and taking only 23s vs 121s TpE. Similarly on Tiny-Imagenet: comparing our Sandwich-small vs SLL-medium, ours has much better clean and robust accuracy, despite having 10M parameters vs 314M, and taking 60s vs 363s TpE.

\begin{table*}[h]
\centering
\caption{Certified robustness of SLL and sandwich model on CIFAR-100 and Tiny-ImageNet datasets. Different from the previous experiment setup on empirical robustness, here we remove the input data normalization and add the last layer normalization. Results of the SLL models are from \cite{araujo2023unified}. Results of the sandwich model are averaged of 3 experiments.}\label{tab:cert-robust}
\vskip 0.15in
\begin{center}
\begin{small}
\begin{sc}
    \begin{tabular}{llrrrrrrr}
    \toprule
    \multicolumn{1}{c}{\multirow{2}[2]{*}{\textbf{Datasets}}} & \multicolumn{1}{c}{\multirow{2}[2]{*}{\textbf{Models}}} & \multicolumn{1}{c}{\multirow{2}[2]{*}{\textbf{\makecell{\textbf{Clean} \\ \textbf{Acc.}}}}} & \multicolumn{3}{c}{\boldmath{}\textbf{Certified Acc. ($\varepsilon$)}\unboldmath{}} & \multicolumn{1}{c}{\multirow{2}[2]{*}{\textbf{\makecell{\textbf{Number of} \\ \textbf{Parameters}}}}} \\
   \cmidrule{4-6}
     & &   & $\frac{36}{255}$ & $\frac{72}{255}$ & $\frac{108}{255}$  \\
    \midrule
    \multirow{2}[4]{*}{CIFAR100}
      & \textbf{SLL Small}  & 44.9 & 34.7 & 26.8 & 20.9 &  41M  \\
      & \textbf{SLL XLarge} & \textbf{46.5} & \textbf{36.5} & \textbf{28.4} & \textbf{22.7} &  236M  \\
      & \textbf{Sandwich}   & 46.3 & 35.3 & 26.3 & 20.3 &  26M  \\
    \midrule
    \multirow{2}[4]{*}{TinyImageNet}
      & \textbf{SLL Small}          & 26.6 & 19.5 & 12.2 & 10.4 & 167M  \\
      & \textbf{SLL X-Large}         & 32.1 & 23.0 & 16.9 & 12.3 & 1.1B  \\
      & \textbf{Sandwich}     & \textbf{33.4} & \textbf{24.7} & \textbf{18.1} & \textbf{13.4} & 39M  \\
    \bottomrule
    \end{tabular}%
\end{sc}
\end{small}
\end{center}
\vskip -0.1in
\end{table*}%

\paragraph{Certified robustness on CIFAR-100 and Tiny-Imagenet.}
We also compare the certified robustness to the SLL approach which outperforms most existing 1-Lipschitz networks \cite{araujo2023unified}. From \cref{tab:cert-robust} we can see that on CIFAR-100, our Sandwich model performs similarly to somewhat larger SLL models (41M parameters vs 26M, i.e. 60\% larger). However it is outperformed by much larger SLL models (236M parameters, 9 times larger than ours). 

On Tiny-Imagenet, however, we see that our model uniformly outperforms SLL models, even the extra large SLL model with 1.1B parameters (28 times larger than ours). Furthermore, our advantage over the four-times larger ``Small'' SLL model is substantial, e.g. 24.7\% vs 19.5\% certified accuracy for $\epsilon= 36/255$. 

%%%%%%%%%%%%%%%%%%%%%%%%%%%%%%%%%%%%%%%%%%%%%%%%%%%%%%%%%%%%%%%%%%%%%%%%%%%%%%%
\section{Conclusions}

In this paper we have introduced a direct parameterization of neural networks that automatically satisfy the SDP-based Lipschitz bounds of \cite{fazlyab2019efficient}. It is a \emph{complete} parameterization, i.e. it can represent all such neural networks. Direct parameterization enables learning of Lipschitz-bounded networks with standard first-order gradient methods, avoiding the need for complex projections or barrier evaluations.  The new parameterization can also be interpreted as a new layer type, the \emph{sandwich layer}. Experiments in robust image classification with both fully-connected and convolutional networks showed that our method outperforms existing models in terms of both empirical and certified accuracy.

% Authors are kindly asked to make their submissions as accessible as possible for everyone including people with disabilities and sensory or neurological differences.
% Tips of how to achieve this and what to pay attention to will be provided on the conference website \url{http://icml.cc/}.

% \section*{Software and Data}

% If a paper is accepted, we strongly encourage the publication of software and data with the
% camera-ready version of the paper whenever appropriate. This can be
% done by including a URL in the camera-ready copy. However, \textbf{do not}
% include URLs that reveal your institution or identity in your
% submission for review. Instead, provide an anonymous URL or upload
% the material as ``Supplementary Material'' into the CMT reviewing
% system. Note that reviewers are not required to look at this material
% when writing their review.

\section*{Acknowledgements}

% \textbf{Do not} include acknowledgements in the initial version of
% the paper submitted for blind review.

This work was partially supported by the Australian Research Council and the NSW Defence Innovation Network.


% \clearpage

\bibliography{ref}

\bibliographystyle{icml2023}

%%%%%%%%%%%%%%%%%%%%%%%%%%%%%%%%%%%%%%%%%%%%%%%%%%%%%%%%%%%%%%%%%%%%%%%%%%%%%%%
%%%%%%%%%%%%%%%%%%%%%%%%%%%%%%%%%%%%%%%%%%%%%%%%%%%%%%%%%%%%%%%%%%%%%%%%%%%%%%%
% APPENDIX
%%%%%%%%%%%%%%%%%%%%%%%%%%%%%%%%%%%%%%%%%%%%%%%%%%%%%%%%%%%%%%%%%%%%%%%%%%%%%%%
%%%%%%%%%%%%%%%%%%%%%%%%%%%%%%%%%%%%%%%%%%%%%%%%%%%%%%%%%%%%%%%%%%%%%%%%%%%%%%%

\appendix
\onecolumn
%%%%%%%%%%%%%%%%%%%%%%%%%%%%%%%%%%%%%%%%%%%%%%%%%%%%%%%%%%%%%%%%%%%%%%%%%%%%%%%
\section{Preliminaries on SDP-based Lipschitz bound}\label{sec:lmi}

We review the theoretical work of SDP-based Lipschitz bound estimation for neural networks from \cite{fazlyab2019efficient,revay2020lipschitz}. Consider an $L$-layer feed-forward network $y=f(x)$ described by the following recursive equation:
\begin{equation}\label{eq:forward-net}
      \begin{split}
        z_0&=x, \\
	z_{k+1} &=\sigma(W_k z_k+b_k),\quad k=0,\ldots,L-1, \\
	y&=W_L z_L+b_L,
  \end{split}
\end{equation}
where $x\in \R^{n_0}, z_k \in \R^{n_k}, y\in \R^{n_{L+1}}$ are the network input, hidden unit of the $k$th layer and network output, respectively. We stack all hidden unit $z_1,\ldots,z_L$ together and obtain a compact form of \eqref{eq:forward-net} as follows:
\begin{equation}
    \begin{split}
    \overset{z}{\overbrace{
    \begin{bmatrix}
        z_1 \\ z_2 \\ \vdots \\ z_L
    \end{bmatrix}}}
    &=\sigma\left(
    \overset{W}{\overbrace{
    \begin{bmatrix}
    0 & & &\\
    W_1 & \ddots & & \\
    \vdots & \ddots & 0 & \\
    0 & \cdots & W_{L-1} & 0
    \end{bmatrix} }}
    \begin{bmatrix}
        z_1 \\ z_2 \\ \vdots \\ z_L
    \end{bmatrix}+
    \overset{U}{\overbrace{
    \begin{bmatrix}
     W_0 \\ 0 \\ \vdots \\ 0
  \end{bmatrix}}}x+
  \overset{b_z}{\overbrace{
  \begin{bmatrix}
     b_0 \\ b_1 \\ \vdots \\ b_{L-1}
  \end{bmatrix}}}\right), \\
  y&=
  \overset{Y}{\overbrace{
  \begin{bmatrix}
    0 & \cdots & 0 & W_L
  \end{bmatrix}}}
  \begin{bmatrix}
        z_1 \\ z_2 \\ \vdots \\ z_L
    \end{bmatrix}+\overset{b_y}{\overbrace{ b_L}}.
    \end{split}
\end{equation}
By letting $z:=\mathrm{col}(z_1,\ldots,z_L)$ and $v:=Wz+Ux+b_z$, we can rewrite the above equation by 
\begin{equation}\label{eq:compact}
    v=Wz+Ux+b_z,\quad z=\sigma(v),\quad y=Yz+b_y.
\end{equation}
Now we introduce the incremental quadratic constraint (iQC) \cite{megretski1997system} for analyzing the activation layer.
\begin{lemma}\label{lem:iqc}
    If Assumption~\ref{asmp:sigma} holds, then for any $\Lambda \in \D_{++}^n$ the following iQC holds for any pair of $(v^a,z^a)$ and $(v^b, z^b)$ satisfying $z=\sigma(v)$:
    \begin{equation}\label{eq:iqc}
        \begin{bmatrix}
            \Delta v^\top \\ \Delta z^\top
        \end{bmatrix}^\top 
        \begin{bmatrix}
            0 & \Lambda \\
            \Lambda & -2\Lambda
        \end{bmatrix}
        \begin{bmatrix}
            \Delta v\\ \Delta z
        \end{bmatrix}\geq 0
    \end{equation}
    where $\Delta v=v^b-v^a$ and $\Delta z=z^b-z^a$.
\end{lemma}
\begin{remark}\label{rem:error}
Assumption~\ref{asmp:sigma} implies that each channel satisfies $2\Delta z_i (\Delta v_i-\Delta z_i) \geq 0$, which can be leads to \eqref{eq:iqc} by a linear conic combination of each channel with multiplier $\Lambda\in \D_{++}^n$. In \cite{fazlyab2019efficient} it was claimed that iQC \eqref{eq:iqc} holds with a richer (more powerful) class of multipliers (i.e. $\Lambda$ is a symmetric matrix), which were previously introduced for robust stability analysis of systems with repeated nonlinearities  \citep{chu1999bounds,damato2001new,kulkarni2002all}. However this is not true: a counterexample was given in \cite{pauli2021training}. Here we give a brief explanation: even if the nonlinearities $\sigma(v_i)$ are repeated when considered as functions of $v_i$, their increments $\Delta z_i=\sigma(v_i^a+\Delta v_i)-\sigma(v_i^a)$ are not repeated when considered as functions of $\Delta v_i$, since $\delta z_i$ depend on the particular $v_i^a$ which generally differs between units.
\end{remark}

\begin{theorem}\label{thm:lmi}
    The feed-forward neural network \eqref{eq:forward-net} is $\gamma$-Lipschitz if Assumption~\ref{asmp:sigma} holds, and there exist an $\Lambda \in \D_{++}^n$ satisfying the following LMI:
    \begin{equation}\label{eq:lmi-2}
        H:=\begin{bmatrix}
        \gamma I & -U^\top \Lambda & 0\\
        -\Lambda U & 2\Lambda-\Lambda W-W^\top \Lambda & -Y^\top \\
        0 & -Y & \gamma I
    \end{bmatrix}\succeq 0.
    \end{equation}
\end{theorem}
\begin{remark}
    In \cite{revay2020lipschitz}, the above LMI condition also applies to more general network structures with full weight matrix $W$. An equivalent form of \eqref{eq:lmi-2} was applied in \cite{fazlyab2019efficient} for a tight Lipschitz bound estimation:
    \begin{equation}
        \min_{\gamma,\Lambda}\; \gamma \quad \mathrm{s.t.}\quad \eqref{eq:lmi-2}
    \end{equation}
    which can be solved by convex programming for moderate models, e.g., $n< 10\mathrm{K}$ in \cite{fazlyab2019efficient}. 
\end{remark}

%%%%%%%%%%%%%%%%%%%%%%%%%%%%%%%%%%%%%%%%%%%%%%%%%%%%%%%%%%%%%%%%%%%%%%%%%%%%%%%
\section{1-Lipschitz convolutional layer}\label{sec:conv}

Our proposed layer parameterization can also incorporate more structured linear operators such as convolution. Let $\hin\in \R^{p\times s\times s}$ be a $p$-channel image tensor with $s\times s$ spatial domain and $\hout \in \R^{q\times s\times s}$ be $q$-channel output tensor. We also let $A\in \R^{q\times q \times s\times s}$ denote a multi-channel convolution operator and similarly for $B\in \R^{q\times p\times s\times s}$. For the sake of simplicity, we assume that the convolutional operators $A,B$ are circular and unstrided. Such assumption can be easily related to plain and/or $2$-strided convolutions, see \cite{trockman2021orthogonalizing}. Similar to \eqref{eq:sandwich-layer}, the proposed convolutional layer can be rewritten as
\begin{equation}\label{eq:conv-layer}
    \vecc({h}_{\mathsf{out}})=\sqrt{2}\mathcal{C}_A^\top \Psi_s\sigma\bigl(\sqrt{2}\Psi_s^{-1}\mathcal{C}_B\vecc(\hin)+b\bigr)
\end{equation}
where $\mathcal{C}_A\in\R^{qs^2\times qs^2},\mathcal{C}_B\in\R^{qs^2\times ps^2}$ are the doubly-circular matrix representations of $A$ and $B$, respectively. For instance, $\vecc(B\ast \hin)=\mathcal{C}_B\vecc(\hin)$ where $\ast$ is the convolution operator. We choose $\Psi_s=\Psi\otimes I_s$ with $\Psi=\diag(e^{d})$ so that individual channel has a constant scaling factor. To ensure that \eqref{eq:conv-layer} is 1-Lipschitz, we need to construct $\mathcal{C}_A,\mathcal{C}_B$ using the Cayley transformation \eqref{eq:cayley}, which involves inverting a highly-structured large matrix $I+\mathcal{C}_Z\in \R^{qs^2\times qs^2}$.

Thanks to the doubly-circular structure, we can perform efficient computation on the Fourier domain. Taking a 2D case for example,  circular convolution of two matrices is simply the elementwise product of their representations in the Fourier domain \cite{jain1989fundamentals}. In \cite{trockman2021orthogonalizing}, the 2D convolution theorem was extended to multi-channel circular convolutions of tensors, which are reduced to a batch of complex matrix-vector products in the Fourier domain rather than elementwise products. For example, the Fourier-domain output related to the $(i,j)^{\mathrm{th}}$ pixel is a matrix-vector product:
\[
\FFT(B\ast \hin)[:,i,j]=\tilde{B}[:,:,i,j]\tilde{h}_{\mathsf{in}}[:,i,j].
\]
where $\tilde{B}[:,:,i,j]\in \C^{q\times p}$ and $\tilde{h}_{\mathsf{in}}[:,i,j]\in \C^{p}$. Here $\mathbb{C}$ denotes the set of complex numbers and $\tilde{x}=\FFT(x)$ is the fast Fourier transformation (FFT) of a multi-channel tensor $x\in\R^{c_1\times \cdots\times c_r\times s\times s}$:
\begin{equation*}
    \FFT(x)[i_1,\ldots,i_r, :,:]=\mathcal{F}_s x[i_1, \ldots, i_r,:,:]\mathcal{F}_s^*
\end{equation*}
where $\mathcal{F}_s[i,j]=\frac{1}{s}e^{-2\pi(i-1)(j-1)\iota/s}$ with $\iota=\sqrt{-1}$. 
Moreover, transposing or inverting a convolution is equivalent to applying the complex version of the same operation to its Fourier domain representation -- a batch of small complex matrices:
\[
\begin{split}
    \FFT(A^\top)[:,:,i,j]=\tilde{A}[:,:,i,j]^{*},\quad \FFT((I+Z)^{-1})[:,:,i,j] = (I+\tilde{Z}[:,:,i,j])^{-1}.
\end{split}
\]
Since the FFT of a real tensor is Hermitian-symmetric, the batch size can be reduced to $s\times (\lfloor s/2\rfloor+1)$. 

%%%%%%%%%%%%%%%%%%%%%%%%%%%%%%%%%%%%%%%%%%%%%%%%%%%%%%%%%%%%%%%%%%%%%%%%%%%%%%%
\section{Weighted Spectral Norm Bounds}\label{sec:norm-bound}

The generalized Clake Jacobian operator of feedforward network $f_\phi$ in \eqref{eq:DNN} has the following form
\[
\mathbf{J}^cf_\phi=W_L \prod_{k=1}^{L} J_{L-k} W_{L-k}  \in \R^{n_{L+1}\times n_0}
\]
where $J_{k}=\mathbf{J}^c\sigma(W_{k}z_{k}+b_k)\in \J_+^{n_{k+1}}$ with $\J_+^{n_{k+1}}$ defined as follows
\begin{equation}\label{eq:Jq}
    \J_{+}^n:=\{\text{diagonal }J\in \R^{n\times n}\mid J_{ii}\in[0,1],\, \forall 1\leq i\leq n\}.
\end{equation}
To learn an $1$-Lipschitz DNN, one can impose the constraints $\|W_k\|\leq 1$ for $k=0,\ldots,L$, i.e., $f_\phi$ satisfies the following spectral norm bound  
\begin{equation}
    \|\mathbf{J}^cf_\phi\|\leq \prod_{k=0}^{L}\|W_k\|\leq 1.
\end{equation}
However, such bound is often quite loose, see an example in \cref{fig:bound}. 

For our proposed model parameterization, we can also estimate the Lipschitz bound via the production of layerwise Lipschitz bounds, i.e.,
\begin{equation}\label{eq:lip-bound}
    \|\mathbf{J}^cf_\phi\|\leq\sqrt{\gamma}\times  \prod_{k=0}^{L-1}\|\mathbf{J}^c s(h_k)\| \times \sqrt{\gamma}\|B_L\|\leq \gamma
\end{equation}
where $s$ is the $1$-Lipschitz sandwich layer function defined in \eqref{eq:sandwich-layer} and $\|B_L\|\leq 1$ by construction. In the following proposition, we show that the layerwise bound in \eqref{eq:lip-bound} is equivalent to weight spectral norm bounds on the weights $W_k$.
\begin{proposition}\label{prop:lbdn-spec}
    The feedforward network \eqref{eq:DNN} with weights \eqref{eq:W-lbdn} satisfies the weighted spectral norm bounds as follows:
    \begin{equation}
        \begin{cases}
        \left\|\frac{1}{\sqrt{2}}B_0^+\Psi_0W_0\right\|\leq \sqrt{\gamma}, \\
        \left\|\frac{1}{2}B_k^\top \Psi_k W_k\Psi_{k-1}^{-1}\bigl(A_{k-1}^\top\bigr)^+\right\|\leq 1,\quad 1\leq k<L,\\
        \left\|\frac{1}{\sqrt{2}}W_L\Psi_{L-1}^{-1}\bigl(A_{L-1}^\top\bigr)^+\right\|\leq \sqrt{\gamma},
    \end{cases}
    \end{equation}
    Moreover, the network is $\gamma$-Lipschitz since
    \begin{equation}\label{eq:weight-bound}
        \|\mathbf{J}^cf_\phi\|\leq \left\|\frac{1}{\sqrt{2}}B_0^+\Psi_0W_0\right\|\times\prod_{k=1}^{L}\left\|\frac{1}{2}B_k^\top \Psi_k W_k\Psi_{k-1}^{-1}\bigl(A_{k-1}^\top\bigr)^+\right\|\times \left\|\frac{1}{\sqrt{2}}W_L\Psi_{L-1}^{-1}\bigl(A_{L-1}^\top\bigr)^+\right\|\leq \gamma.
    \end{equation}
\end{proposition} 
\begin{remark}
    For 1-Lipschitz DNNs, our model parameterization allows for the spectral norm bounds of both individual layer and the whole network to be larger than 1, while the network Lipschitz constant is still bounded by a weighted layerwise spectral bound of 1, see the example in \cref{fig:bound}.
\end{remark}

%%%%%%%%%%%%%%%%%%%%%%%%%%%%%%%%%%%%%%%%%%%%%%%%%%%%%%%%%%%%%%%%%%%%%%%%%%%%%%%
\section{Proofs}\label{sec:proof}

%%----------------------------------------------------------------
\subsection{Proof of \cref{lem:iqc}}
Given any pair of $(v^a,z^a)$ and $(v^b, z^b)$ satisfying $z=\sigma(v)$, we have $\Delta z = \sigma(v^b)-\sigma(v^a):=J^{ab}\Delta v$ with $\Delta z=z^b-z^a$ and $\Delta v=v^b-v^a$, where $J^{ab}\in \J_+^q$ with $\J_+^q$ defined in \eqref{eq:Jq}. Therefore, we can have
\[
\begin{bmatrix}
            \Delta v^\top \\ \Delta z^\top
        \end{bmatrix}^\top 
        \begin{bmatrix}
            0 & \Lambda \\
            \Lambda & -2\Lambda
        \end{bmatrix}
        \begin{bmatrix}
            \Delta v\\ \Delta z
        \end{bmatrix}=2\Delta z^\top \Lambda (\Delta v- \Delta z)=2\Delta v^\top J^{ab}\Lambda(I-J^{ab})\Delta v\geq 0.
\]

%%----------------------------------------------------------------
\subsection{Proof of \cref{thm:lmi}}
We first apply Schur complement to \eqref{eq:lmi-2}, which yields
\begin{equation*}
    \begin{bmatrix}
        \gamma I & -U^\top \Lambda \\
        -\Lambda U & 2\Lambda-\Lambda W-W^\top \Lambda -\frac{1}{\gamma}Y^\top Y
    \end{bmatrix}\succ 0.
\end{equation*}
Then, by left-multiplying the above equation by $\begin{bmatrix}\Delta x^\top & \Delta z^\top \end{bmatrix}$ and right-multiplying $\begin{bmatrix}\Delta x^\top & \Delta z^\top \end{bmatrix}^\top$ we can obtain
\begin{equation}
    \gamma \|\Delta x\|^2-\frac{1}{\gamma}\|\Delta y\|^2-2\Delta z^\top \Lambda \Delta z-2\Delta z^\top \Lambda (W\Delta z+U\Delta x)=\gamma \|\Delta x\|^2-\frac{1}{\gamma}\|\Delta y\|^2-2\Delta z^\top \Lambda (\Delta z-\Delta v)\geq 0,
\end{equation}
which further implies that \eqref{eq:forward-net} is $\gamma$-Lipschitz since
\[
\gamma \|\Delta x\|^2-\frac{1}{\gamma}\|\Delta y\|^2\geq 2\Delta z^\top \Lambda (\Delta v-\Delta z)\geq 0
\]
where the last inequality follows by Lemma~\ref{lem:iqc}.

\subsection{Proof of \cref{thm:lbdn}}
\paragraph{Sufficient.} We  show that \eqref{eq:lmi-2} holds with $\Lambda=\diag(\Lambda_0,\ldots,\Lambda_{L-1})$ where $\Lambda_k=\Psi_k^2$. Since the block structure of $H$ is a chordal graph, $H\succeq 0$ is equivalent to the existence of a chordal decomposition \cite{zheng2021chordal}:
\begin{equation}\label{eq:chordal}
    H=\sum_{k=0}^L E_k H_k E_k^\top
\end{equation}
where $0\preceq H_k\in \R^{(n_k+n_{k+1})\times (n_k+n_{k+1})} $ and $E_k=\begin{bmatrix}\mathbf{0}_{a,k} & \mathbf{I}_{b,k} & \mathbf{0}_{c,k} \end{bmatrix}$ with $\mathbf{I}_{b,k}$ being the identity matrix the same size as $H_k$, and $\mathbf{0}_{a,k}, \mathbf{0}_{c,k}$ being zero matrices of appropriate dimension. We then construct $H_k$ as follows.

For $k=0$, we take
\begin{equation}\label{eq:H0}
    H_0=\begin{bmatrix}
        \gamma I & -\sqrt{2\gamma} B_0^\top \Psi_0\\
        -\sqrt{2\gamma}\Psi_0 B_0 & 2\Psi_0(I-A_0A_0^\top) \Psi_0
    \end{bmatrix}.
\end{equation}
Note that $H_0\succeq 0$ since $[H_0]_{11}=\gamma I\succ 0$, and the Schur complement to $[H_0]_{11}$  yields
\[
2\Psi_0(I-A_0A_0^\top) \Psi_0-\sqrt{2\gamma}\Psi_0B_0 \frac{1}{\gamma} I \sqrt{2\gamma}B_0^\top \Psi_0=2\Psi_0(I-A_0A_0^\top-B_0B_0^\top)\Psi_0=0.
\]
For $k=1,\ldots,L-1$ we take 
\begin{equation}\label{eq:Hk}
    H_k=\begin{bmatrix}
        2\Psi_{k-1} A_{k-1} A_{k-1}^\top \Psi_{k-1} & -2\Psi_{k-1} A_{k-1}B_k^\top \Psi_k \\
        -2\Psi_k B_k A_{k-1}^\top \Psi_{k-1} & 2\Psi_{k} (I-A_{k} A_{k}^\top) \Psi_{k}
    \end{bmatrix}.
\end{equation}
If $A_{k-1}$ is zero, then it is trival to have $H_k\succeq 0$. For nonzero $A_{k-1}$, we can verify that $H_k\succeq 0$ since the Schur complement to $[H_k]_{11}$ shows
\[
\begin{split}
    & 2\Psi_{k} (I-A_{k} A_{k}^\top) \Psi_{k} - 2\Psi_k B_k A_{k-1}^\top \Psi_{k-1} \left(2\Psi_{k-1} A_{k-1} A_{k-1}^\top \Psi_{k-1}\right)^{+} 2\Psi_{k-1} A_{k-1}B_k^\top \Psi_k \\
    =& 2\Psi_k(I-A_kA_k^\top-B_kB_k^\top)\Psi_k+2\Psi_kB_k(I-A_{k-1}^+A_{k-1})B_k^\top \Psi_k \\
    =& 2\Psi_kB_k(I-A_{k-1}^+A_{k-1})B_k^\top \Psi_k\succeq 0
\end{split}
\]
where $X^{+}$ denotes the Moore–Penrose inverse of the matrix $X$, and it satisfies $I-X^+X\succeq 0$.

For $k=L$ we take
\begin{equation}\label{eq:HL}
    H_L=\begin{bmatrix}
        2\Psi_{L-1} A_{L-1}A_{L-1}^\top \Psi_{L-1} & -\sqrt{2\gamma} A_{L-1} B_L^\top \Psi_{L-1} \\
        -\sqrt{2\gamma}\Psi_{L-1} B_LA_{L-1}^\top  &\gamma I
    \end{bmatrix}.
\end{equation}
Similarly, we can conclude $H_L\succeq 0$ using Schur complement
\[
\gamma I - \sqrt{2\gamma}\Psi_{L-1} B_LA_{L-1}^\top \left( 2\Psi_{L-1} A_{L-1}A_{L-1}^\top \Psi_{L-1} \right)^{+} \sqrt{2\gamma} A_{L-1} B_L^\top \Psi_{L-1}=\gamma \Psi_{L-1} B_L (I-A_{L-1}^+ A_{L-1})B_L^\top \Psi_{L-1}\succeq 0.
\]
We now show that $H_k$ with $k=0,\ldots L$ satisfy the chordal decomposition \eqref{eq:chordal} holds since
\begin{gather*}
    [H_k]_{21}=-2\Psi_k B_k A_{k-1}^\top \Psi_{k-1}=-\Psi_{k}^2(2\Psi_k^{-1}B_k A_{k-1}^\top \Psi_{k-1}) =-\Lambda_k W_k, \\
    [H_k]_{22}+[H_{k+1}]_{11}=2\Psi_{k} (I-A_{k} A_{k}^\top) \Psi_{k}+2\Psi_{k} A_{k} A_{k}^\top \Psi_{k}=2\Psi_k^2=2\Lambda_k.
\end{gather*}
Finally, we conclude that $H\succeq 0$ from \cite{zheng2021chordal}[Theorem 2.1].

\paragraph{Necessary.} For any $W_k$ and $\Lambda_k$ satisfying \eqref{eq:lmi-2}, we will find set of free variables $d_k,X_k,Y_k$ such that \eqref{eq:W-lbdn} holds. We take $\Psi_k=\Lambda^{\frac{1}{2}}$ which further leads to $d_k=\diag(\log \Psi_k)$. By letting $A_{-1}=I,\Psi_{-1}=\sqrt{\gamma/2}I$ and $\Psi_L=\sqrt{2/\gamma}I$ we then construct $A_k,B_k$ recursively via
\begin{equation}
    B_k=\frac{1}{2}\Psi_kW_k\Psi_{k-1}^{-1}A_{k-1}^{-\top},\quad A_k=\mathrm{chol}(I-B_kB_k^\top)Q_k
\end{equation}
where $\mathrm{chol}(\cdot)$ denotes the Cholesky factorization, $Q_k$ is an arbitrary orthogonal matrix such that $A_k$ does not have eigenvalue of $-1$. If $A_{k-1}$ is non-invertible but non-zero, we replace $A_{k-1}^{-\top}$ with $\bigl(A_{k-1}^{+}\bigr)^\top$. If $A_{k-1}=0$ (i.e. $W_k=0$), we simply reset $A_{k-1}=I$. It is easy to verify that $\Psi_k,A_k$ and $B_k$ satisfy the model parameterization \eqref{eq:W-lbdn}. Finally, we can construct $X_k,Y_k$ using \eqref{eq:XYZ}, which is well-defined as $A_k$ does not have eigenvalue of $-1$.

%%----------------------------------------------------------------
\subsection{Proof of \cref{thm:layer}}
The proposed layer \eqref{eq:sandwich-layer} can be rewritten as a compact network \eqref{eq:compact} with $W=0$, $ Y=\sqrt{2}A^\top \Psi$ and $U=\sqrt{2}\Psi^{-1}B$, i.e.,
\[
v=U\hin+b,\quad z=\sigma(v),\quad \hout = Y z.
\]
From the model parameterization \eqref{eq:cayley} we have $A A^\top+B B^\top=I$, which further implies
\begin{equation*}
    2\Psi^2-Y^\top Y-\Psi^2 UU^\top \Psi^2= 2\Psi^2-2\Psi A A^\top \Psi-2\Psi B B^\top \Psi=2\Psi(I-A A^\top-B B^\top)\Psi=0
\end{equation*}
By applying Schur complement twice to the above equation we have
\[
\begin{bmatrix}
         I & -U^\top \Psi^2 & 0\\
        -\Psi^2 U & 2\Psi^2 & -Y^\top \\
        0 & -Y &  I
    \end{bmatrix}\succeq 0.
\]
Then, the $1$-Lipschitzness of \eqref{eq:sandwich-layer} is obtained by \cref{thm:lmi}.
\subsection{Proof of \cref{prop:linear-layer}}
\paragraph{Sufficient.} It is a direct corollary of \cref{thm:layer} by taking the identity operator as the nonlinear activation.

\paragraph{Necessary.} Here we give a constructive proof. That is, given a weight matrix $W$ with $\|W\|\leq 1$, we will find a (generally non-unique) pair of $(X,Y)$ such that $2A^\top B=W$ with $A,B$ given by \eqref{eq:cayley}.

We first construct $A,B$ from $W$. Since it is obvious for $W=0$, we consider the case with nonzero $W$. First, we take a singular value decomposition (SVD) of $W$, i.e. $W=U_w\Sigma_wV_w^\top$ where $U_w$ is a $q\times q$ orthogonal matrix, $\Sigma_w$ is an $q\times p$ rectangular diagonal matrix with $\Sigma_{w,ii}\geq 0 $ non-increasing, $V_w$ is a $p\times p$ orthogonal matrix. Then, we consider the candidates for $A$ and $B$ as follows:
\begin{equation}\label{eq:AB}
    A=U\Sigma_a U_w^\top,\quad B=U\Sigma_b V_w^\top
\end{equation}
where $\Sigma_a$ is a diagonal matrix, $\Sigma_b$ a rectangular diagonal matrix $U\in\R^{q\times q}$ an orthogonal matrix. By substituting \eqref{eq:AB} into the equalities $AA^\top+BB^\top=I_q$ and $W=2A^\top B$ we have 
\begin{equation}\label{eq:Sigma}
    \begin{split}
        \Sigma_a^2+\Sigma_{b'}^2=I_q, \quad
        2\Sigma_a\Sigma_{b'}=\Sigma_{w'}
    \end{split}
\end{equation}
where $\Sigma_{b'},\Sigma_{w'}\in \R^{q\times q}$ are obtained by either removing the extra columns of zeros on the right or adding extra rows of zeros at the bottom to $\Sigma_b$ and $\Sigma_w$, respectively. The solution to \eqref{eq:Sigma} is 
\begin{equation}
    \Sigma_{a,ii}=\frac{1}{2}\left(\sqrt{1+\Sigma_{w',ii}}+\sqrt{1-\Sigma_{w',ii}}\right),\quad \Sigma_{b',ii}=\frac{1}{2}\left(\sqrt{1+\Sigma_{w',ii}}-\sqrt{1-\Sigma_{w',ii}}\right)
\end{equation}
where are well-defined as $\|W\|\leq 1$. Now we can obtain $\Sigma_b$ from $\Sigma_{b'}$ by removing extra rows of zeros at the bottom or adding extra columns of zeros on the right. At last, we pick up any orthogonal matrix $U$ such that $A=U\Sigma_a U_w^\top$ does not have eigenvalue of $-1$.

The next step is to find a pair of $(X,Y)$ such that 
\begin{equation}
    A^\top = (I+Z)^{-1}(I-Z), \quad B^\top=-2Y(I+Z)^{-1},\quad Z=X-X^\top+Y^\top Y.
\end{equation}
One solution to the above equation is 
\begin{equation}\label{eq:XYZ}
    Z=(I-A^\top)(I+A^\top)^{-1},\quad Y=-\frac{1}{2} B^\top (I+Z), \quad X=\frac{1}{2} \mathrm{tril}(Z-Z^\top)
\end{equation}
where $\mathrm{tril}(W)$ denotes the strictly lower triangle part of $W$.
Note that the above solution is well-defined since $A$ does not has eigenvalue of $-1$. 

\subsection{Proof of \cref{prop:lbdn-spec}}
From \eqref{eq:H0} we have 
\[
H_0=\begin{bmatrix}
    \gamma I & -W_0^\top\Psi_0^2 \\
    -\Psi_0^2 W_0 & 2\Psi_0 B_0 B_0^\top \Psi_0
\end{bmatrix}\succeq 0.
\]
Applying the Schur complement yields $\gamma I-1/2W_0^\top \Psi_0 (B_0B_0^\top)^+ \Psi_0 W_0\succeq 0$, which implies $\|B_0^+\Psi_0 W_0\|\leq \sqrt{2\gamma}$. From \eqref{eq:Hk} we obtain
\[
\begin{split}
    &H_k=\begin{bmatrix}
        2\Psi_{k-1} A_{k-1} A_{k-1}^\top \Psi_{k-1} & -W_k^\top \Psi_k^2 \\
        -\Psi_k^2 W_k & 2\Psi_{k} B_kB_k^\top \Psi_{k}
    \end{bmatrix}\succeq 0 \\
    \Rightarrow & \Psi_{k-1} A_{k-1} A_{k-1}^\top \Psi_{k-1} - \frac{1}{4}W_k^\top \Psi_k (B_kB_k^\top)^+\Psi_k W_k\succeq 0 \\
    \Rightarrow & I-\frac{1}{4} A_{k-1}^+\Psi_{k-1}^{-\top} W_k^\top \Psi_k (B_kB_k^\top)^+\Psi_k W_k \Psi_{k-1}^{-1} \bigl(A_{k-1}^\top\bigr)^+\succeq 0 \\
    \Rightarrow & \left\|\frac{1}{2} B_k^+ \Psi_k W_k\Psi_{k-1}^{-1} \bigl(A_{k-1}^\top\bigr)^+\right\|\leq 1.
\end{split}
\]
Similarly, from \eqref{eq:HL} we have
\[
\begin{split}
    &H_L=\begin{bmatrix}
        2\Psi_{L-1} A_{L-1}A_{L-1}^\top \Psi_{L-1} & -W_L^\top \\
        -W_L &\gamma I
    \end{bmatrix}\succeq 0 \Rightarrow \left\|W_L\Psi_{L-1}^{-1}\bigl(A_{L-1}^\top\bigr)^+\right\|\leq \sqrt{2\gamma}.
\end{split}
\]
The bound of Jacobian operator $\mathbf{J}^c f$ is then obtained by
\[
\begin{split}
    \|\mathbf{J}^c &f\|=\|W_LJ_{L-1}W_{L-1}\cdots J_0 W_0\| \\ 
    =&\left\|\frac{1}{2}W_L\Psi_{L-1}^{-1}\bigl(A_{L-1}^\top\bigr)^+ (2A_{L-1}^\top J_{L-1} B_{L-1}) \prod_{k=L-1}^{1}\left(\frac{1}{2} B_k^+ \Psi_k W_k\Psi_{k-1}^{-1} \bigl(A_{k-1}^\top\bigr)^+ \right)(2A_{k-1}^\top J_{k-1} B_{k-1})(B_0^+\Psi_0 W_0)\right\|\\
    \leq &\left\|\frac{1}{\sqrt{2}}B_0^+\Psi_0W_0\right\|\times\prod_{k=1}^{L}\left\|\frac{1}{2}B_k^\top \Psi_k W_k\Psi_{k-1}^{-1}\bigl(A_{k-1}^\top\bigr)^+\right\|\times \left\|\frac{1}{\sqrt{2}}W_L\Psi_{L-1}^{-1}\bigl(A_{L-1}^\top\bigr)^+\right\| 
    \leq \gamma
\end{split}
\]
where the first inequality follows as $2A_{k}^\top J_{k} B_{k}$ is the Clake Jacobian of a 1-Lipschitz layer \eqref{eq:sandwich-layer}, i.e. $\|2A_{k}^\top J_{k} B_{k}\|\leq 1$.

%%%%%%%%%%%%%%%%%%%%%%%%%%%%%%%%%%%%%%%%%%%%%%%%%%%%%%%%%%%%%%%%%%%%%%%%%%%%%%%
\section{Training details}\label{sec:training-details}

For all experiments, we used a piecewise triangular learning rate  \cite{coleman2017dawnbench} with maximum rate of $0.01$. We use Adam \cite{kingma2014adam} and ReLU as our default optimizaer and activation, respectively. Because the Cayley transform in \eqref{eq:cayley} involves both linear and quadratic terms,  we implemented the weight normalization method from \cite{winston2020monotone}. That is, we reparameterize $X,Y$ in $Z=X-X^\top + Y^\top Y$ by $g\frac{X}{\|X\|_F}$ and $h \frac{Y}{\|Y\|_F}$ with learable scalars $g,h$. We search for the empirical lower Lipschitz bound $\underline{\gamma}$ of a network $f_\theta$ by a PGD-like method, i.e., updating the input $x$ and its deviation $\delta_x$ based on the gradient of $\|f_\theta(x+\Delta x)-f_\theta(x)\|/\|\Delta x\|$. As we are interested in the global lower Lipschitz bound, we do not project $x$ and $x+\Delta x$ into any compact region. For image classification tasks, we applied data augmentation used by \cite{araujo2023unified}. All experiments were performed on an Nvidia A5000.  

\paragraph{Toy example.} For the curve fitting experiment, we take 300 and 200 samples $(x_i,y_i)$ with $x_i\sim\mathcal{U}([-2,2])$ for training and testing, respectively. We use batch size of 50 and Lipschitz bounds of 1, 5 and 10. All models for the toy example have 8 hidden layers. We choose width of 128, 128, 128 and 86 for AOL, orthogonal, SLL and sandwich layers, respectively, so that each model size is about 130K. We use MSE loss and train models for 200 epochs.

\begin{table}[t]
\caption{Model architectures for MNIST.  }
\label{tab:mnist-model}
\vskip 0.15in
\begin{center}
\begin{small}
% \begin{sc}
\begin{tabular}{l l l}
	    \toprule
	    MLP & Orthogonal & Sandwich  \\ \midrule
	    \texttt{Fc(784,256)} & \texttt{OgFc(784,256)} & \texttt{SwFc(784,190)}   \\
		  \texttt{Fc(256,256)} & \texttt{OgFc(256,256)}  & \texttt{SwFc(190,190)}   \\ 
            \texttt{Fc(256,128)} & \texttt{OgFc(256,128)} & \texttt{SwFc(190,128)}  \\ 
            \texttt{Lin(128,10)} & \texttt{OgLin(128,10)} & \texttt{SwLin(128,10)} \\
	    \bottomrule
	\end{tabular}
% \end{sc}
\end{small}
\end{center}
\vskip -0.15in
\end{table}

\begin{table}[t]
\caption{Model architectures for CIFAR-10/100 and Tiny-ImageNet. We use $w=1,2,4$ to denote the \emph{small}, \emph{medium} and \emph{large} models. The default kernel size for all convolution is 3.  For orthogonal and sandwich convolution, we use the emulated 2-stride from \cite{trockman2021orthogonalizing} when \texttt{s=2} is indicated. For CNN, \texttt{s=2} refers to the standard 2-stride operation. Since the AOL layer does not support stride operation, we add average pooling at the end to convolution layers. Here \texttt{ncls} denotes the number of classes in the dataset, e.g. 100 for CIFAR-100 and 200 for Tiny-ImageNet.}
\label{tab:emp-model}
\vskip 0.15in
\begin{center}
\begin{small}
% \begin{sc}
\begin{tabular}{l l l l}
	    \toprule
	    CNN & AOL & Orthogonal & Sandwich  \\ \midrule
	    \texttt{Conv(3,32*w)} & \texttt{AolConv(3,32*w)} & \texttt{OgConv(3,32*w)} & \texttt{SwConv(3,32*w)}   \\
		  \texttt{Conv(32*w,32*w,s=2)} & \texttt{AolConv(32*w,32*w)}  & \texttt{OgConv(32*w,32*w,s=2)} & \texttt{SwConv(32*w,32*w,s=2)}   \\ 
            \texttt{Conv(32*w,64*w)} & \texttt{AolConv(32*w,64*w)} & \texttt{OgConv(32*w,64*w)} & \texttt{SwConv(32*w,64*w)}  \\ 
            \texttt{Conv(64*w,64*w,s=2)} & \texttt{AolConv(64*w,64*w)} & \texttt{OgConv(64*w,64*w,s=2)} & \texttt{SwConv(64*w,64*w,s=2)} \\
            \texttt{Flatten}& \texttt{AvgPool(4),Flatten} & \texttt{Flatten} & \texttt{Flatten} \\
            \texttt{Fc(4096*w,640*w} & \texttt{AolFc(4096*w,640*w)} & \texttt{OgFc(4096*w,640*w)} & \texttt{SwFc(4096*w,512*w)} \\
            \texttt{Fc(640*w,512*w)} & \texttt{AolFc(640*w,512*w)} & \texttt{OgFc(640*w,512*w)} & \texttt{SwFc(512*w,512*w)}\\
            \texttt{Lin(512*w,ncls)} & \texttt{AolLin(512*w,ncls)} & \texttt{OgLin(512*w,ncls)} & \texttt{SwLin(512*w,ncls)} \\
	    \bottomrule
	\end{tabular}
% \end{sc}
\end{small}
\end{center}
\vskip -0.2in
\end{table}

\begin{table}[bt]
\caption{Sandwich models in the experiment of certified robustness. Here \texttt{LLN} stands for the Last Layer Normalization \cite{singlaimproved} which can improve the certified robustness when the number of classes become large. }
\label{tab:cert-model}
\vskip 0.15in
\begin{center}
\begin{small}
% \begin{sc}
\begin{tabular}{l l}
	    \toprule
	    CIFAR-100 & TinyImageNet  \\ \midrule
	     \texttt{SwConv(3,64)} & \texttt{SwConv(3,64)}   \\
		\texttt{SwConv(64,64,s=2)} & \texttt{SwConv(64,64,s=2)}   \\ 
            \texttt{SwConv(64,128)}  & \texttt{SwConv(64,128)}  \\ 
             \texttt{SwConv(128,128,s=2)} & \texttt{SwConv(128,128,s=2)} \\
             \texttt{SwConv(128,256)}  & \texttt{SwConv(128,256)}  \\ 
             \texttt{SwConv(256,256,s=2)} & \texttt{SwConv(256,256,s=2)} \\
             -  & \texttt{SwConv(256,512)}  \\ 
             - & \texttt{SwConv(512,512,s=2)} \\
             \texttt{SwFc(1024,2048)} & \texttt{SwFc(2048,2048)} \\
             \texttt{SwFc(2048,2048)} & \texttt{SwFc(2048,2048)}\\
             \texttt{SwFc(2048,1024)} & \texttt{SwFc(2048,1024)}\\
             \texttt{LLN(1024,100)} & \texttt{LLN(1024,200)} \\
	    \bottomrule
	\end{tabular}
% \end{sc}
\end{small}
\end{center}
\vskip -0.2in
\end{table}

\paragraph{Image classification.}
We trained small fully-connected model on MNIST and the KWLarge network from \cite{li2019preventing} on CIFAR-10. To make the different models have similar number of parameters in the same experiment, we slightly reduce the hidden layer width of sandwich model in the MNIST experiment and increases width of the first fully-connected layer of CNN and orthogonal models. The model architectures are reported in \cref{tab:mnist-model} - \ref{tab:emp-model}. We used the same loss function as \cite{trockman2021orthogonalizing} for MNIST and CIFAR-10 datasets. The Lipschitz bounds $\gamma$ are chosen to be 0.1, 0.5, 1.0 for MNIST and 1,10,100 for CIFAR-10. All models are trained with normalized input data for 100 epochs. The data normalization layer increases the Lipschitz bound of the network to $\approx$ $4.1\gamma$.  

For the experiment of empirical robustness, model architectures with different sizes are reported in \cref{tab:emp-model}. The SLL model with small, medium and large size can be found in \cite{araujo2023unified}. We train models with different Lipschitz bounds of $\{0.5,1,2,\ldots,16\}$. We found that $\gamma=2$ for CIFAR-100 and $\gamma=1$ for Tiny-ImageNet achieve the best robust accuracy for the perturbation size of $\epsilon=36/255$. All models are trained with normalized input data for 100 epochs. 

We also compare the certified robustness to the SLL model. Slightly different from the experimental setup for empirical robustness comparison, we remove the data normalization and use the Last Layer Normalization (LLN) proposed by \cite{singlaimproved} which can improve the certified accuracy when the number of classes becomes large. We set the Lipschitz bound of sandwich and SLL models to 1. But the Lipschitz constant of the composited model could be larger than 1 due to LLN  Due to LLN. The certified accuracy is then normalized by the last layer \cite{singlaimproved}. Also, we remove the data normalization for better certified robustness. For all experiments on CIFAR-100 and Tiny-ImageNet, we use the CrossEntropy loss as in \cite{prach2022almost} with temperature of 0.25 and an offset value $3\sqrt{2}/2$ .

\end{document}